\documentclass[twoside]{article}

\usepackage[accepted]{aistats2026}
\usepackage{graphicx}
\usepackage{algorithm}
\usepackage{algpseudocode}
\usepackage{booktabs}
\usepackage{multirow}
\usepackage{amssymb}
\usepackage{amsmath}
\usepackage{amsthm}
\usepackage{newfloat}
\usepackage{listings}
\usepackage{thm-restate}
\usepackage{url}

\theoremstyle{plain}

\usepackage[round]{natbib}

\let\cite\citep

\begin{document}

% If your paper is accepted and the title of your paper is very long,
% the style will print as headings an error message. Use the following
% command to supply a shorter title of your paper so that it can be
% used as headings.
%
\runningtitle{ReTrack: Data Unlearning in Diffusion Models through Redirecting the Denoising Trajectory}

% If your paper is accepted and the number of authors is large, the
% style will print as headings an error message. Use the following
% command to supply a shorter version of the author names so that
% they can be used as headings (for example, use only the surnames)
%
% \runningauthor{Surname 1, Surname 2, Surname 3, ...., Surname n}

\twocolumn[

\aistatstitle{ReTrack: Data Unlearning in Diffusion Models through \\ Redirecting the Denoising Trajectory}

\aistatsauthor{ Qitan Shi \And Cheng Jin \And  Jiawei Zhang \And Yuantao Gu }

\aistatsaddress{ Department of Electronic Engineering \\ Tsinghua University \\ Beijing, China} ]

\begin{abstract}
Diffusion models excel at generating high-quality, diverse images but also suffer from undesirable training data memorization, raising critical privacy and safety concerns. Data unlearning has emerged to mitigate this issue by removing the influence of specific data through fine-tuning rather than retraining from scratch. We propose ReTrack, a fast and effective data unlearning method for diffusion models. ReTrack employs importance sampling to construct a more efficient unbiased fine-tuning loss. This loss is further approximated by retaining only the dominant terms, thereby reducing computational cost. This yields an interpretable objective that redirects denoising trajectories toward the \(k\)-nearest neighbors, enabling efficient unlearning while preserving generative quality. Experiments on MNIST T-Shirt, CelebA-HQ, CIFAR-10, and Stable Diffusion show that ReTrack achieves state-of-the-art performance, striking the best trade-off between unlearning strength and generation quality preservation.
\end{abstract}

\section{INTRODUCTION}

In recent years, diffusion models have brought about a revolutionary breakthrough in image generation. Representative models such as Denoising Diffusion Probabilistic Models \cite{ddpm} and Stable Diffusion \cite{sd} have completely transformed the field of generative artificial intelligence with their exceptional image generation quality and powerful expressive capabilities. These models can generate highly realistic images, demonstrating unprecedented potential in various applications including artistic creation \cite{wu2022creative, zhang2023inversion}, image inpainting \cite{lugmayr2022repaint, manukyan2023hd}, super-resolution \cite{li2022srdiff, saharia2022image} and data augmentation \cite{alimisis2025advances}. 

With their growing capacity, diffusion models have shown a pronounced tendency to memorize training data. Several studies have demonstrated that these models can unintentionally reproduce specific training samples \cite{gu2023memorization, somepalli2023diffusion, carlini2023extracting, chen2024towards}. Given that the training data of such models are typically sourced from heterogeneous web scrapes, user uploads, or open datasets, they often contain potentially harmful, sensitive, or copyrighted content. The memorization of such data raises privacy, legal, and ethical concerns.

In order to eliminate the influence of these data on the model, the ideal approach would be to retrain the model after removing these data from the training dataset. However, this approach incurs prohibitively large computational cost and is often impractical \cite{xu2024machine, wang2024machine}.
A research field aiming to modify a pretrained model so that it behaves as though it had never been trained on some data in the training dataset without retraining the whole model is known as \textit{machine unlearning}. 
Although there have been some early works in the machine unlearning field, most of these works either requires modifications to the pretraining phase \cite{wu2020deltagrad, bourtoule2021machine, graves2021amnesiac} or is specifically designed for supervised machine learning \cite{neggrad, izzo2021approximate, wang2022federated, perifanis2024sftc} and cannot be directly applied to pretrained diffusion models.

To handle this problem, recent works have begun exploring machine unlearning methods specifically designed for diffusion models. These works can be broadly categorized into \textit{concept unlearning} and \textit{data unlearning} \cite{siss}, with this paper focusing on the latter. Unlike concept unlearning which aims to forget harmful or inappropriate abstract concepts in text-conditioning such as ``nudity" and ``violence" \cite{fan2023salun, gandikota2023erasing, gandikota2024unified, zhang2024forget, erasediff}, data unlearning seeks to achieve a sample-level unlearning, i.e., to remove the influence of specific illicit or privacy-sensitive data in the training dataset on the diffusion models, while preserving the influence of other remaining data. 
For data unlearning, there are two relatively intuitive yet limited baseline methods: the vanilla method which directly fine‑tunes the model on the remaining data by applying the standard diffusion loss, and NegGrad \cite{neggrad} which performs gradient ascent on the data to be unlearned.
The former often fails to achieve true forgetting, and the latter usually leads to over-forgetting, causing a serious degradation of model generation quality.
To address these shortcomings, more data unlearning methods have been proposed in recent years, including:
Variational Diffusion Unlearning (VDU) \cite{vdu} which incorporates variational inference techniques into the data unlearning for diffusion models and enables the removal of specific data without having access to the remaining data by approximating the posterior distribution over the parameters of the unlearned model; 
Subtracted Importance Sampled Scores (SISS) \cite{siss} which leverages importance sampling by introducing a mixture distribution to combine the vanilla loss and the NegGrad loss, thereby enabling selective unlearning of specific data while preserving the model’s generation quality.

In this work, we propose \textit{ReTrack}, a data unlearning method for diffusion models by \textbf{\underline{Re}}directing the denoising \textbf{\underline{Tra}}je\textbf{\underline{c}}tory to the \textbf{\underline{k}}-nearest neighbors. 
First, based on the observation that the vanilla fine-tuning loss is unbiased but highly inefficient in sampling, we propose a new loss function derived from the vanilla loss that applies importance sampling to focus the unlearning process on samples that need to be unlearned, thereby greatly accelerating unlearning.
Second, to address the challenge of exactly evaluating the full loss function, we approximate it with the \(k\) most important terms, which correspond to the \(k\)-nearest neighbors of the samples to be unlearned. 
Intuitively, the approximated loss function fine-tunes the model such that denoising trajectories originally directed toward samples to be unlearned are redirected to their \(k\)-nearest neighbors within all the remaining data. Since these neighbors are close in distance to the unlearned samples in data space, the required correction is small, which accelerates training and improves stability. Moreover, redirecting toward real and plausible targets helps maintain the quality of generated outputs during unlearning.
Finally, we conduct experiments on MNIST T-Shirt, CelebA-HQ, CIFAR-10 and Stable Diffusion, and demonstrate that our method achieves the best unlearning results among all existing methods that are able to preserve the model's generation quality, meaning that our method strikes the best trade-off between unlearning strength and generation quality preservation.

Our main contributions can be summarized as follows:
\begin{itemize}
    \item We propose a novel data unlearning fine-tuning loss that preserves unbiasedness while substantially improving sampling efficiency during fine-tuning, enabling more effective unlearning.
    \item We derive a truncated approximation to the proposed loss by retaining only the dominant terms, thereby reducing computational cost. We provide an intuitive explanation for this approximation.
    \item We conduct extensive experiments on MNIST T-Shirt, CelebA-HQ, CIFAR-10, and Stable Diffusion, demonstrating that our method achieves efficient unlearning while maintaining the generative quality of the models. Code is available at \url{https://github.com/sqt24/ReTrack}.
\end{itemize}

\section{PRELIMINARY}

\subsection{Diffusion Models}

Diffusion models are a class of powerful generative frameworks that learn complex data distributions by coupling two complementary processes: a \textit{forward diffusion process}, which gradually perturbs the data with Gaussian noise, and a \textit{reverse denoising process}, which learns to reconstruct the data from noise \cite{ddpm}.

Suppose the training dataset is \(A\) where the data dimension is \(d\). In the forward process, given a training sample \(\boldsymbol{a} \in A\) and a diffusion timestep \(t\in \{1,2,\cdots,T\}\) where \(T\) denotes the total number of diffusion steps, a noisy sample \(\boldsymbol{x}_t\) at timestep \(t\) is produced according to
\begin{equation*}
\boldsymbol{x}_t = \gamma_t \boldsymbol{a} + \sigma_t \boldsymbol{\epsilon},
\end{equation*}
where \(\gamma_t\) and \(\sigma_t\) are time-dependent coefficients specified by a predefined noise schedule, and \(\boldsymbol{\epsilon} \sim \mathcal{N}(\cdot;\boldsymbol{0}, \boldsymbol{\mathrm{I}}_d)\) denotes standard Gaussian noise. Conditioned on the clean input \(\boldsymbol{a}\), the noisy sample \(\boldsymbol{x}_t\) follows
\begin{equation*}
q_t(\boldsymbol{x}_t|\boldsymbol{a}) = \mathcal{N}(\boldsymbol{x}_t; \gamma_t \boldsymbol{a}, \sigma_t^2 \boldsymbol{\mathrm{I}}_d).
\end{equation*}
As \(t\) increases, the noisy sample \(\boldsymbol{x}_t\) contains less and less data and more and more noise, and eventually be completely changed to Gaussian noise.

The reverse process is modeled by a neural network that aims to approximate the conditional expectation of the noise term \(\boldsymbol{\epsilon}\) given the perturbed input \(\boldsymbol{x}_t\) and the timestep \(t\). Formally, the model \({\boldsymbol{\epsilon}}_{\boldsymbol{\theta}}\) parameterized by \(\boldsymbol{\theta}\) is trained to minimize the expected squared error
\begin{equation*}
\mathcal{L}_{\text{train}}(\boldsymbol{\theta}) = \mathbb{E}_{t,\boldsymbol{a}\sim A, \boldsymbol{x}_t\sim q_t(\cdot|\boldsymbol{a})} \left[ \left\| {\boldsymbol{\epsilon}}_{\boldsymbol{\theta}}(\boldsymbol{x}_t, t)-\boldsymbol{\epsilon} \right\|^2 \right],
\end{equation*}
This training objective encourages the model to approximate the true posterior mean \(\mathbb{E}_{\boldsymbol{\epsilon}}[\boldsymbol{\epsilon} | \boldsymbol{x}_t, t]\), which characterizes the reverse process in a variational framework. By training the denoising model \(\boldsymbol{\epsilon}_{\boldsymbol{\theta}}\), the data distribution can be fitted indirectly.

At inference time, sample generation proceeds by initializing with a Gaussian noise \(\boldsymbol{x}_T \sim \mathcal{N}(\cdot;\boldsymbol{0}, \boldsymbol{\mathrm{I}}_d)\) and iteratively applying the learned denoising model according to a time-discretized approximation of the reverse diffusion process, obtaining the denoising trajectory \(\boldsymbol{x}_T, \boldsymbol{x}_{T-1}, \cdots, \boldsymbol{x}_1, \boldsymbol{x}_0\). This process gradually transforms Gaussian noise \(\boldsymbol{x}_T\) into a data sample \(\boldsymbol{x}_0\) following the learned data distribution, thereby enabling data generation.

\subsection{Data Unlearning for Diffusion Models}

\subsubsection{Definition}

The data unlearning problem is mainly studied for unconditional diffusion models, which can be formulated as follows: given a training dataset \(A\) and a diffusion model \(\boldsymbol{\epsilon}_{\boldsymbol{\theta}}\) pretrained on \(A\), our goal is to fine-tune the model with relatively modest computational cost so that it forgets the influence of a subset \(A_u\subset A\) (referred to as the \textit{unlearning set}), while preserving the influence of the remaining data \(A_r=A\setminus A_u\) (referred to as the \textit{remaining set}). 
In other words, we seek to obtain a fine-tuned model \(\boldsymbol{\epsilon}_{\boldsymbol{\theta}^*}\) with comparatively low computational cost, such that it behaves as though it had been pretrained solely on \(A_r\).

\subsubsection{Prior Methods}

\paragraph{Vanilla}
Vanilla serves as a baseline method that simply fine-tune the pretrained model only on remaining set \(A_r\) using the standard diffusion loss:
\[
\mathcal{L}_{\text{vanilla}}(\boldsymbol{\theta}) = \mathbb{E}_{t,\boldsymbol{a}_r\sim A_r,\boldsymbol{x}_t\sim q_t(\cdot|\boldsymbol{a}_r)}\left[\left\|\boldsymbol{\epsilon}_{\boldsymbol{\theta}}(\boldsymbol{x}_t,t)-\boldsymbol{\epsilon}\right\|_2^2\right].
\]
While intuitive, this method completely ignores any information about the unlearning set \(A_u\). As a result, even after many fine-tuning steps, it often fails to effectively remove the influence of \(A_u\).

\paragraph{NegGrad}
NegGrad \cite{neggrad} directly performs gradient ascent on samples from the unlearning set \(A_u\) to push the model’s predictions away from the true noise estimations, which is equivalent to performing gradient descent on
\[
\mathcal{L}_{\text{NegGrad}}(\boldsymbol{\theta}) = -\mathbb{E}_{t,\boldsymbol{a}_u\sim A_u,\boldsymbol{x}_t\sim q_t(\cdot|\boldsymbol{a}_u)}\left[\|\boldsymbol{\epsilon}_{\boldsymbol{\theta}}(\boldsymbol{x}_t,t)-\boldsymbol{\epsilon}\|_2^2\right].
\]
However, NegGrad cannot guarantee that the model’s performance on the remaining set \(A_r\) is preserved. It often leads to over-forgetting, whereby the model also forgets \(A_r\) while attempting to forget \(A_u\).

\paragraph{EraseDiff}
Although EraseDiff \cite{erasediff} is originally designed for concept unlearning, following the setting described by \citet{siss}, it can also be applied to data unlearning task by guiding the model’s predictions on \(A_u\) toward pure noise:
\begin{align*}
&\mathcal{L}_{\text{EraseDiff}}(\boldsymbol{\theta}) = \mathbb{E}_{t,\boldsymbol{a}_r\sim A_r,\boldsymbol{x}_t\sim q_t(\cdot|\boldsymbol{a}_r)}\left[\|\boldsymbol{\epsilon}_{\boldsymbol{\theta}}(\boldsymbol{x}_t,t)-\boldsymbol{\epsilon}\|_2^2\right] \\&\hspace{0em}+ \lambda \mathbb{E}_{t,\boldsymbol{a}_u\sim A_u,\boldsymbol{x}_t\sim q_t(\cdot|\boldsymbol{a}_u),\boldsymbol{\epsilon}_u \sim\mathcal{U}[0,1]^d}\left[\|\boldsymbol{\epsilon}_{\boldsymbol{\theta}}(\boldsymbol{x}_t,t)-\boldsymbol{\epsilon}_u\|_2^2\right],
\end{align*}
where the first term is used to preserve \(A_r\), and the second term is used to unlearn \(A_u\) by pushing the model's prediction toward pure noise.
By solving an optimization problem on \(\lambda\), EraseDiff ensures that the overall gradient descent direction of the loss function simultaneously serves as the descent direction for both of its constituent terms, thereby forgetting \(A_u\) while preserving \(A_r\).

\paragraph{SISS}
SISS \cite{siss} strikes a balance between the vanilla fine-tuning loss and the NegGrad loss, and improves computational efficiency by employing importance sampling:
\begin{align*}
&\hspace{0em}\mathcal{L}_{\text{SISS}}(\boldsymbol{\theta}) = \mathbb{E}_{t,\boldsymbol{a}_r\sim A_r,\boldsymbol{a}_u\sim A_u,\boldsymbol{x}_t\sim q_t(\cdot|\boldsymbol{a}_r,\boldsymbol{a}_u)} \\&\hspace{2em} \left[ \frac{q_t(\boldsymbol{x}_t|\boldsymbol{a}_r)}{q_t(\boldsymbol{x}_t|\boldsymbol{a}_r,\boldsymbol{a}_u)}\left\|\boldsymbol{\epsilon}_{\boldsymbol{\theta}}(\boldsymbol{x}_t,t)-\frac{\boldsymbol{x}_t-\gamma_t \boldsymbol{a}_r}{\sigma_t}\right\|_2^2  \right.\\&\hspace{4em}\left. -s\frac{q_t(\boldsymbol{x}_t|\boldsymbol{a}_u)}{q_t(\boldsymbol{x}_t|\boldsymbol{a}_r,\boldsymbol{a}_u)}\left\|\boldsymbol{\epsilon}_{\boldsymbol{\theta}}(\boldsymbol{x}_t,t)-\frac{\boldsymbol{x}_t-\gamma_t \boldsymbol{a}_u}{\sigma_t}\right\|_2^2\right],
\end{align*}
where the first term corresponds to the vanilla fine-tuning loss, the second term corresponds to the NegGrad loss, and
\[
q_t(\boldsymbol{x}_t|\boldsymbol{a}_r,\boldsymbol{a}_u) = (1-\lambda)q_t(\boldsymbol{x}_t|\boldsymbol{a}_r) + \lambda q_t(\boldsymbol{x}_t|\boldsymbol{a}_u)
\]
is a mixture distribution between \(q_t(\boldsymbol{x}_t|\boldsymbol{a}_r)\) and \(q_t(\boldsymbol{x}_t|\boldsymbol{a}_u)\) parameterized by \(\lambda\in [0,1]\), and \(s>0\) is another hyperparameter controlling the strength of unlearning.

\paragraph{VDU}
VDU \cite{vdu} performs data unlearning by introducing a variational Bayesian framework. However, it relies on access to multiple independently pretrained diffusion models on the same dataset, which is impractical under our experimental settings. Therefore, we exclude it from our subsequent comparisons.

\section{METHODOLOGY}

\subsection{Motivation}

We first review vanilla fine-tuning method, which directly performs denoising matching training on the remaining dataset, with the corresponding loss function given by
\[
\mathcal{L}_{\text{vanilla}}(\boldsymbol{\theta}) = \mathbb{E}_{t,\boldsymbol{a}_r\sim A_r,\boldsymbol{x}_t\sim q_t(\cdot|\boldsymbol{a}_r)}\left[\|\boldsymbol{\epsilon}_{\boldsymbol{\theta}}(\boldsymbol{x}_t,t)-\boldsymbol{\epsilon}\|_2^2\right].
\]
If one were to train a model from scratch with this objective, the resulting model would converge to the ideal solution trained only on \(A_r\). Thus, \(\mathcal{L}_{\mathrm{vanilla}}\) is an \emph{unbiased} fine-tuning objective, a property that is important for stability and for preserving generative quality after unlearning.

However, \(\mathcal{L}_{\mathrm{vanilla}}\) is highly \emph{inefficient} for unlearning. Samples \(\boldsymbol{x}_t\) are always generated from \(\boldsymbol{a}_r\in A_r\), so training concentrates on regions surrounding \(A_r\). In contrast, the regions where unlearning must occur are precisely the neighborhoods of \(A_u\). As a result, the fine-tuned model’s behavior near \(A_u\) remains largely unchanged, and the influence of \(A_u\) persists even after many steps of fine-tuning.
This observation motivates us to improve upon the vanilla method by enhancing sampling efficiency, while retaining its favorable property of unbiasedness.

\subsection{Importance-weighted Unlearning Objective}

Inspired by the idea of importance sampling, we modify the sampling distribution of \(\mathcal{L}_{\text{vanilla}}\) by shifting the sampling of \(\boldsymbol{x}_t\) from regions concentrated around \(A_r\) to regions concentrated around \(A_u\) while maintaining the unbiasedness of the fine-tuning loss. This modification improves the sampling efficiency of the objective during fine-tuning and thus enables faster unlearning. The proposed fine-tuning loss is
\begin{align*}
\mathcal{L}_{\text{unlearn}}(\boldsymbol{\theta})
&= \mathbb{E}_{t,\boldsymbol{a}_u\sim A_u,\boldsymbol{x}_t\sim q_t(\cdot|\boldsymbol{a}_u)}  \\&\hspace{-2em} \left[\sum_{\boldsymbol{a}_r\in A_r} w_t(\boldsymbol{x}_t|\boldsymbol{a}_r)\left\|\boldsymbol{\epsilon}_{\boldsymbol{\theta}}(\boldsymbol{x}_t,t)-\frac{\boldsymbol{x}_t-\gamma_t \boldsymbol{a}_r}{\sigma_t}\right\|_2^2\right],
\end{align*}
where the weight term
\[
w_t(\boldsymbol{x}_t|\boldsymbol{a}_r) = \frac{q_t(\boldsymbol{x}_t|\boldsymbol{a}_r)}{\sum_{\boldsymbol{a}_r'\in A_r}q_t(\boldsymbol{x}_t|\boldsymbol{a}_r')}
\]
is introduced by the importance sampling procedure. Note that, via importance sampling, we change the sampling distribution of \(\boldsymbol{x}_t\) from \(\boldsymbol{x}_t \sim q_t(\cdot | \boldsymbol{a}_r)\) to \(\boldsymbol{x}_t \sim q_t(\cdot | \boldsymbol{a}_u)\). The unbiasedness of \(\mathcal{L}_{\text{unlearn}}\) is guaranteed by the following proposition:
\begin{restatable}{proposition}{equivalence}\label{prop:equivalence}
The two fine-tuning loss functions \(\mathcal{L}_{\textup{vanilla}}\) and \(\mathcal{L}_{\textup{unlearn}}\) are equivalent.
\end{restatable}
The proof is provided in the appendix.
Although the two loss functions \(\mathcal{L}_{\text{vanilla}}\) and \(\mathcal{L}_{\text{unlearn}}\) are theoretically equivalent, in practice, when approximating the expectation via sampling, our method achieves better unlearning performance within a limited number of fine-tuning steps by optimizing the sampling strategy to focus more on the regions around samples in the unlearning set \(A_u\) instead of the remaining set \(A_r\).
More importantly, different from some \emph{biased} loss function terms employed in previous work like gradient ascent term in \(\mathcal{L}_{\text{NegGrad}}\) and \(\mathcal{L}_{\text{SISS}}\) or pure noise guidance term in \(\mathcal{L}_{\text{EraseDiff}}\), \(\mathcal{L}_{\text{unlearn}}\) is an unbiased loss. This property is crucial for preserving model stability and avoiding the quality degradation commonly observed with biased methods.

\subsection{Tractable Approximation with Intuitive Interpretation}

Although theoretically valid, directly evaluating \(\mathcal{L}_{\text{unlearn}}\) can be time-consuming. For a given timestep \(t\), a datum \(\boldsymbol{a}_u\) sampled from the unlearning set \(A_u\), and the corresponding noisy sample \(\boldsymbol{x}_t \sim q_t(\cdot | \boldsymbol{a}_u)\), it requires computing a weighted average over the entire remaining set \(A_r\) with weights \(w_t(\boldsymbol{x}_t|\boldsymbol{a}_r) \propto q_t(\boldsymbol{x}_t | \boldsymbol{a}_r)\). When the size of \(A_r\) is large, this becomes computationally expensive, which motivates approximation.

Note that
\[
q_t(\boldsymbol{x}_t | \boldsymbol{a}_r)
\propto \exp\left(-\frac{\lVert \boldsymbol{x}_t - \gamma_t \boldsymbol{a}_r \rVert_2^2}{2\sigma_t^2}\right),
\]
therefore, as \(\gamma_t \boldsymbol{a}_r\) moves away from \(\boldsymbol{x}_t\), the likelihoods \(q_t(\boldsymbol{x}_t | \boldsymbol{a}_r)\) decay exponentially with the squared Euclidean distance \(\| \boldsymbol{x}_t - \gamma_t \boldsymbol{a}_r \|_2^2\). Consequently, in the weighted average most terms receive nearly negligible weights, and the value is determined primarily by a small subset with the largest weights. It is therefore natural to truncate and estimate the full weighted average by computing it only over the \(k\) items with the largest weights.

In fact, the \(k\) elements in \(A_r\) that maximize the weights on average can be identified by the following proposition:
\begin{restatable}{proposition}{maxk}\label{prop:maxk}
Given \(\boldsymbol{a}_u\) sampled from the unlearning set \(A_u\), the \(k\) samples in the remaining set \(A_r\) with the largest expected likelihoods 
\[
\mathbb{E}_{t,\boldsymbol{x}_t\sim q_t(\cdot|\boldsymbol{a}_u)}\left[q_t(\boldsymbol{x}_t|\boldsymbol{a}_r)\right]
\] 
are the \(k\)-nearest neighbors of \(\boldsymbol{a}_u\) under the Euclidean metric.
\end{restatable}
The proof is provided in the appendix. Based on this proposition, we truncate the summation in the original \(\mathcal{L}_{\text{unlearn}}\) and approximate the full sum using only the largest \(k\) terms, which yields the following approximate fine-tuning loss
\begin{align*}
&\mathcal{L}_{\text{unlearn}}(\boldsymbol{\theta};k) = \mathbb{E}_{t,\boldsymbol{a}_u\sim A_u,\boldsymbol{x}_t\sim q_t(\cdot|\boldsymbol{a}_u)}  \\&\hspace{1.5em} \left[\sum_{\boldsymbol{a}_r\in \mathcal{S}_{k}(\boldsymbol{a}_u)} \tilde{w}_t(\boldsymbol{x}_t|\boldsymbol{a}_r)\left\|\boldsymbol{\epsilon}_{\boldsymbol{\theta}}(\boldsymbol{x}_t,t)-\frac{\boldsymbol{x}_t-\gamma_t \boldsymbol{a}_r}{\sigma_t}\right\|_2^2\right],
\end{align*}
where the truncated weights are defined as
\[
\tilde{w}_t(\boldsymbol{x}_t|\boldsymbol{a}_r) = \frac{q_t(\boldsymbol{x}_t|\boldsymbol{a}_r)}{\sum_{\boldsymbol{a}_r'\in \mathcal{S}_{k}(\boldsymbol{a}_u)}q_t(\boldsymbol{x}_t|\boldsymbol{a}_r')},
\]
and 
\(\mathcal{S}_{k}(\boldsymbol{a}_u)\)
denotes the set of \(\boldsymbol{a}_u\)'s \(k\)-nearest neighbors in \(A_r\).
This can be viewed as computing a weighted average only over the \(k\)-nearest neighbors of \(\boldsymbol{a}_u\) using the truncated weights \(\tilde{w}_t(\boldsymbol{x}_t|\boldsymbol{a}_r)\). Owing to the exponential decay of the likelihood, in practice the choice of \(k\) can be much smaller than the size of \(A_r\), which substantially simplifies the evaluation of the loss function.

\begin{figure}[t]
\centering
\includegraphics[width=0.98\columnwidth]{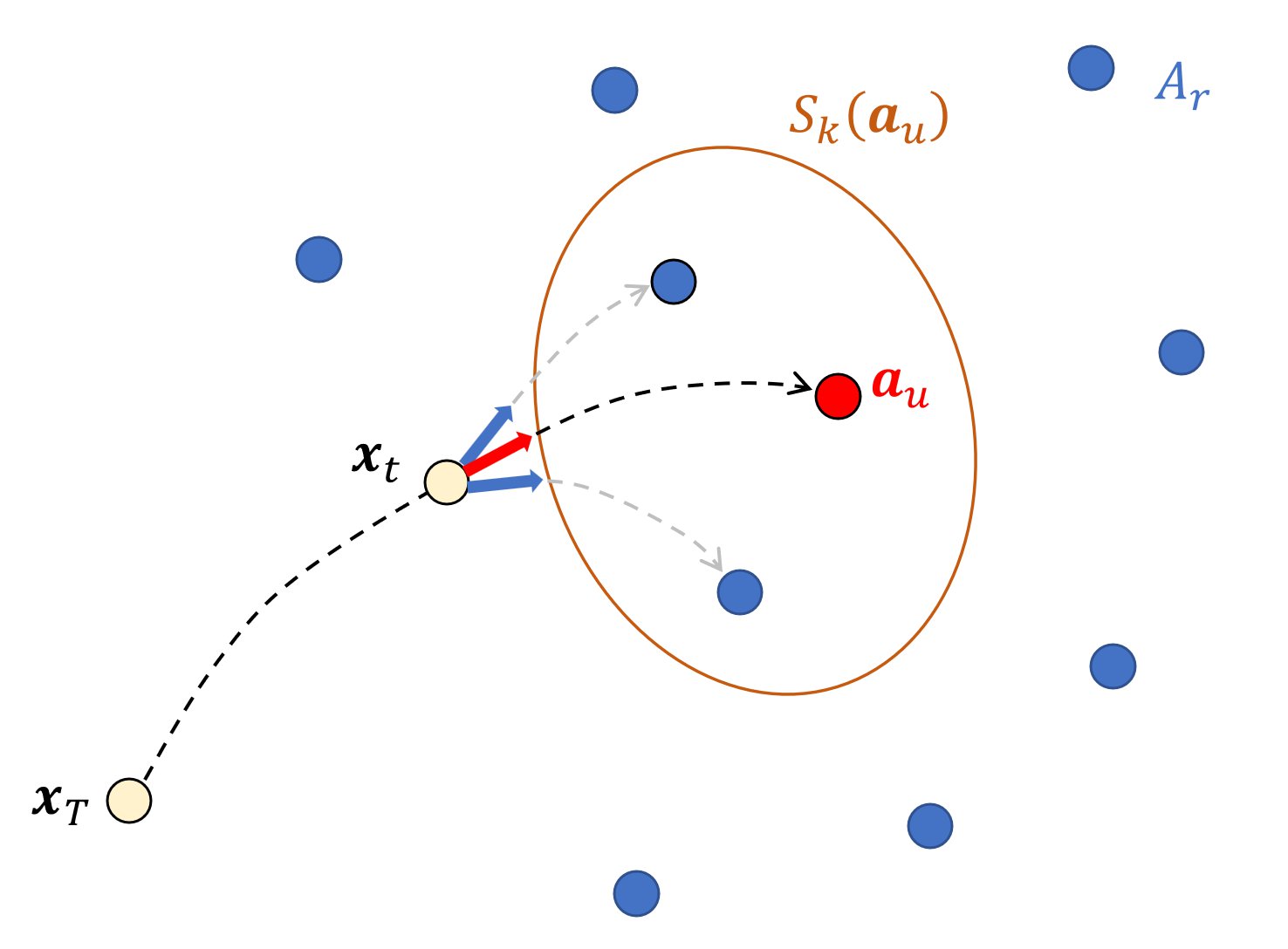}
\caption{
An intuitive schematic of our method. By changing the diffusion model's noise prediction on \(\boldsymbol{x}_t\), 
the denoising trajectory that originally pointed to the unlearning sample \(\boldsymbol{a}_u\) is redirected to its \(k\)-nearest neighbors, thus achieving fast unlearning while preserving generation quality.
}
\label{pic:method}
\end{figure}

We provide an intuitive analysis of the approximate fine-tuning loss function. During fine-tuning, our method can be viewed as adding noise to \(\boldsymbol{a}_u\) to obtain the noisy sample \(\boldsymbol{x}_t\), and then training the model to denoise it toward \(\boldsymbol{a}_r \in \mathcal{S}_{k}(\boldsymbol{a}_u)\) as if \(\boldsymbol{x}_t\) had been produced by adding noise to \(\boldsymbol{a}_r\). For a pretrained model, since information about \(\boldsymbol{a}_u\) is already encoded internally, denoising the \(\boldsymbol{x}_t\) obtained from \(\boldsymbol{a}_u\) will typically drive the denoising trajectory \(\boldsymbol{x}_t,\boldsymbol{x}_{t-1},\cdots,\boldsymbol{x}_0\) back toward \(\boldsymbol{a}_u\), especially when the noise level is small. However, this fine-tuning loss forces the model to redirect the denoising trajectory that would converge to \(\boldsymbol{a}_u\) toward its nearest neighbors in the remaining set \(A_r\), thereby preventing the fine-tuned model from generating undesirable results and achieving efficient unlearning. Figure \ref{pic:method} provides an intuitive illustration of the redirection. Because \(\boldsymbol{a}_r \in \mathcal{S}_{k}(\boldsymbol{a}_u)\) is close to \(\boldsymbol{a}_u\), the correction required in the model output is small, which accelerates the fine-tuning procedure and improves training stability. Moreover, by steering the denoising trajectory toward other real data, the method maintains reasonable outputs during unlearning and thus better mitigates degradation in generation quality. This preservation of realism is consistent with the unbiasedness mentioned in our earlier theoretical analysis.

\subsection{Final Objective}

Finally, due to the finite capacity of practical networks, fine-tuning only in regions around the unlearning set \(A_u\) can cause catastrophic forgetting elsewhere. To mitigate this issue, we introduce the vanilla fine-tuning loss as a regularizer to avoid over-forgetting on the remaining set \(A_r\). The final loss of our ReTrack method is formulated as an interpolation between the unlearning term and the regularization term:
\[
\mathcal{L}_{\text{ReTrack}}(\boldsymbol{\theta}) = \lambda\,\mathcal{L}_{\text{unlearn}}(\boldsymbol{\theta};k) + (1-\lambda)\,\mathcal{L}_{\text{vanilla}}(\boldsymbol{\theta}),
\]
where \(\lambda \in [0,1]\) is a hyperparameter that controls the strength of unlearning. The complete fine-tuning process, implemented with stochastic gradient descent, is shown in Algorithm \ref{alg:knn}. Since the unlearning set \(A_u\) and the remaining set \(A_r\) are fixed throughout fine-tuning, the \(k\)-nearest neighbors can be computed once in advance and stored, and there is no need to recompute at each iteration. Moreover, although the unlearning term sums over \(k\) neighbors, these terms share the same network prediction \(\boldsymbol{\epsilon}_{\boldsymbol{\theta}}\), and thus require only a single function evaluation. As a result, the overall computational cost of our method remains comparable to that of prior methods.

\begin{algorithm}[tb]
\caption{The fine-tuning process of ReTrack.}
\label{alg:knn}
\textbf{Input}: Pretrained model \(\boldsymbol{\epsilon}_{\boldsymbol{\theta}}\), unlearning set \(A_u\), remaining set \(A_r\), fine-tuning steps \(N\).\\
\textbf{Output}: Unlearned model \(\boldsymbol{\epsilon}_{\boldsymbol{\theta}^*}\).

\begin{algorithmic}[1]
\State For each \(\boldsymbol{a}_u\) in \(A_u\), find its \(k\)-nearest neighbors in \(A_r\) under the Euclidean metric.
\For{\(n=1\) \textbf{to} \(N\)}
\State Sample a timestep \(t\in\{1,2,\cdots,T\}\).
\State Sample \(\boldsymbol{a}_u\) from \(A_u\), \(\boldsymbol{x}_t\sim q_t(\cdot|\boldsymbol{a}_u)\) and compute 
\Statex \resizebox{\linewidth}{!}{\(
\mathcal{L}_1(\boldsymbol{\theta}) = \sum_{\boldsymbol{a}_r\in \mathcal{S}_{k}(\boldsymbol{a}_u)} \tilde{w}_t(\boldsymbol{x}_t|\boldsymbol{a}_r) \left\|\boldsymbol{\epsilon}_{\boldsymbol{\theta}}(\boldsymbol{x}_t,t)-\frac{\boldsymbol{x}_t-\gamma_t \boldsymbol{a}_r}{\sigma_t}\right\|_2^2.
\)}
\State Sample \(\boldsymbol{a}_r\) from \(A_r\), \(\boldsymbol{x}_t\sim q_t(\cdot|\boldsymbol{a}_r)\) and compute \(\mathcal{L}_2(\boldsymbol{\theta}) = \left\|\boldsymbol{\epsilon}_{\boldsymbol{\theta}}(\boldsymbol{x}_t,t)-\boldsymbol{\epsilon}\right\|_2^2.\)
\State Compute \(\mathcal{L}(\boldsymbol{\theta}) = 
\lambda\mathcal{L}_1(\boldsymbol{\theta}) + (1-\lambda)\mathcal{L}_2(\boldsymbol{\theta})\).
\State  Take gradient descent step on \(\nabla_{\boldsymbol{\theta}} \mathcal{L}(\boldsymbol{\theta})\).
\EndFor
\end{algorithmic}
\end{algorithm}

\section{EXPERIMENTS}

\subsection{Experimental Setups}

\subsubsection{Datasets.}
\begin{figure}[t]
\centering
\includegraphics[width=0.99\columnwidth]{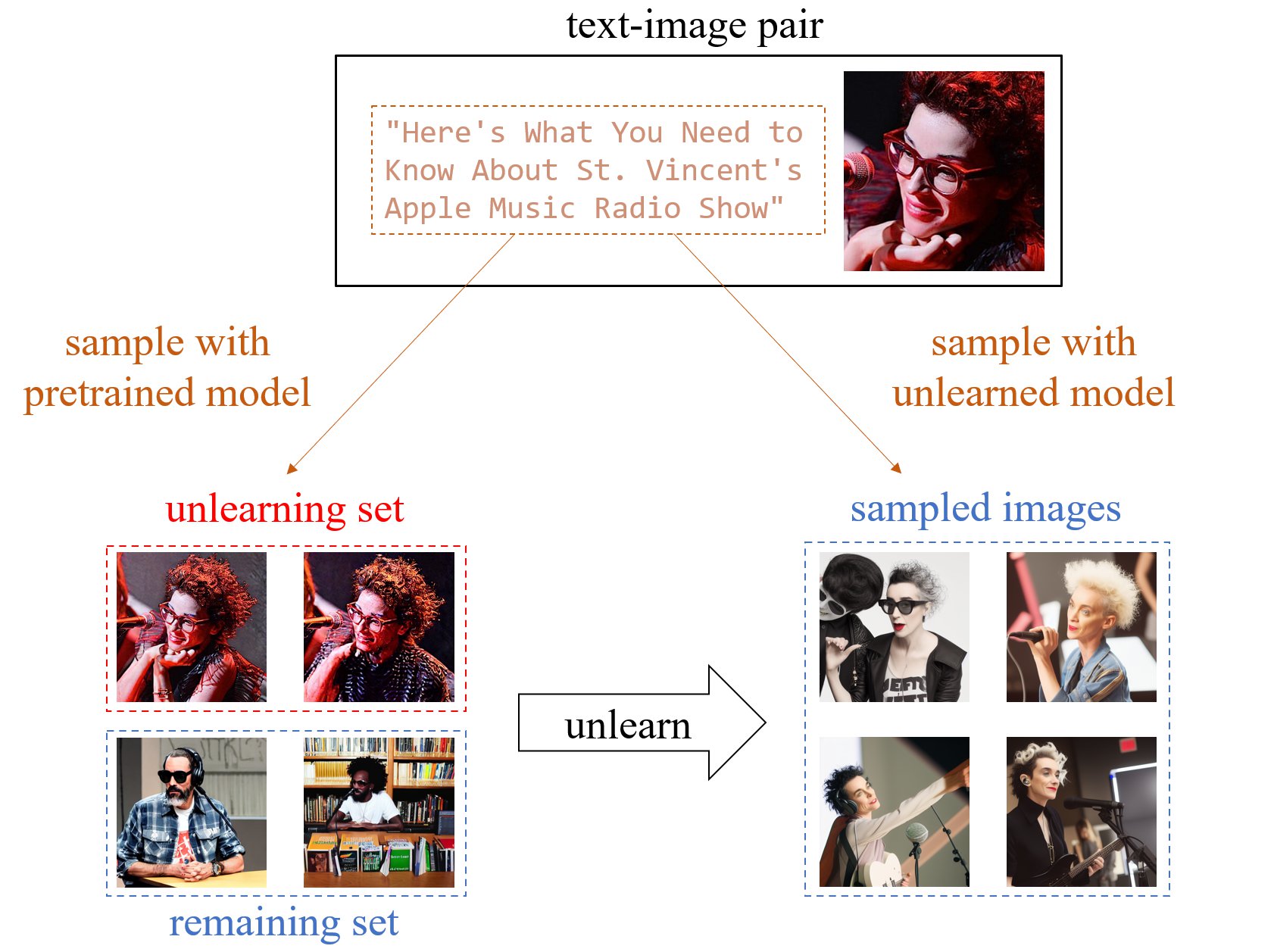}
\caption{
Construction of the Stable Diffusion dataset. For each text-image pair, in order to forget the corresponding image in this text condition, the pretrained model is used to sample images in the given text condition and clustered to construct the unlearning set and remaining set required by the data unlearning method. After unlearning, the model no longer generates the corresponding image in the same text condition.
}
\label{pic:sd_dataset}
\end{figure}

We conduct experiments on four datasets: MNIST T-Shirt, CelebA-HQ, CIFAR-10, and Stable Diffusion. 
The MNIST T-Shirt dataset is a relatively simple test case constructed by \citet{siss} by augmenting the 60,000 handwritten digit images of the MNIST dataset \cite{mnist} with T-Shirt images drawn from Fashion-MNIST dataset \cite{fashion_mnist} at a 1\% ratio, and the goal of unlearning is to make the model forget the T-Shirt images. 
For CelebA-HQ dataset \cite{celeba_hq}, we use the 256×256 resolution version that comprises 30,000 high quality facial images. 
For CIFAR-10 \cite{cifar10}, we use the training split of 50,000 images.
The unlearning dataset for the Stable Diffusion is also constructed by \citet{siss}.
By fixing the model’s text‐conditioned input, Stable Diffusion can be treated as an unconditional generative model for a specific concept, and thus can be employed to evaluate data unlearning methods.
For each text‑image pair appearing in the training dataset of Stable Diffusion, unlearning set and remaining set are required in order to perform data unlearning methods so that the model forgets the original image in the training dataset in this text condition. Therefore, 128 images are sampled using the prompt and then clustered. Those images sufficiently similar to the original training images are designated as the unlearning set, while the remainder formed the remaining set, as demonstrated in Figure \ref{pic:sd_dataset}.

\subsubsection{Evaluation metrics.}
To measure whether an unlearning method can effectively forget images in the unlearning set while preserving generation quality, we evaluate each method using both quality metrics and unlearning metrics.

For quality metrics, we use the standard FID \cite{fid}, IS \cite{is}, and CLIP-IQA \cite{clip_iqa} metrics to measure generation quality of the unlearned model.

\begin{figure}[t]
\centering
\includegraphics[width=0.98\columnwidth]{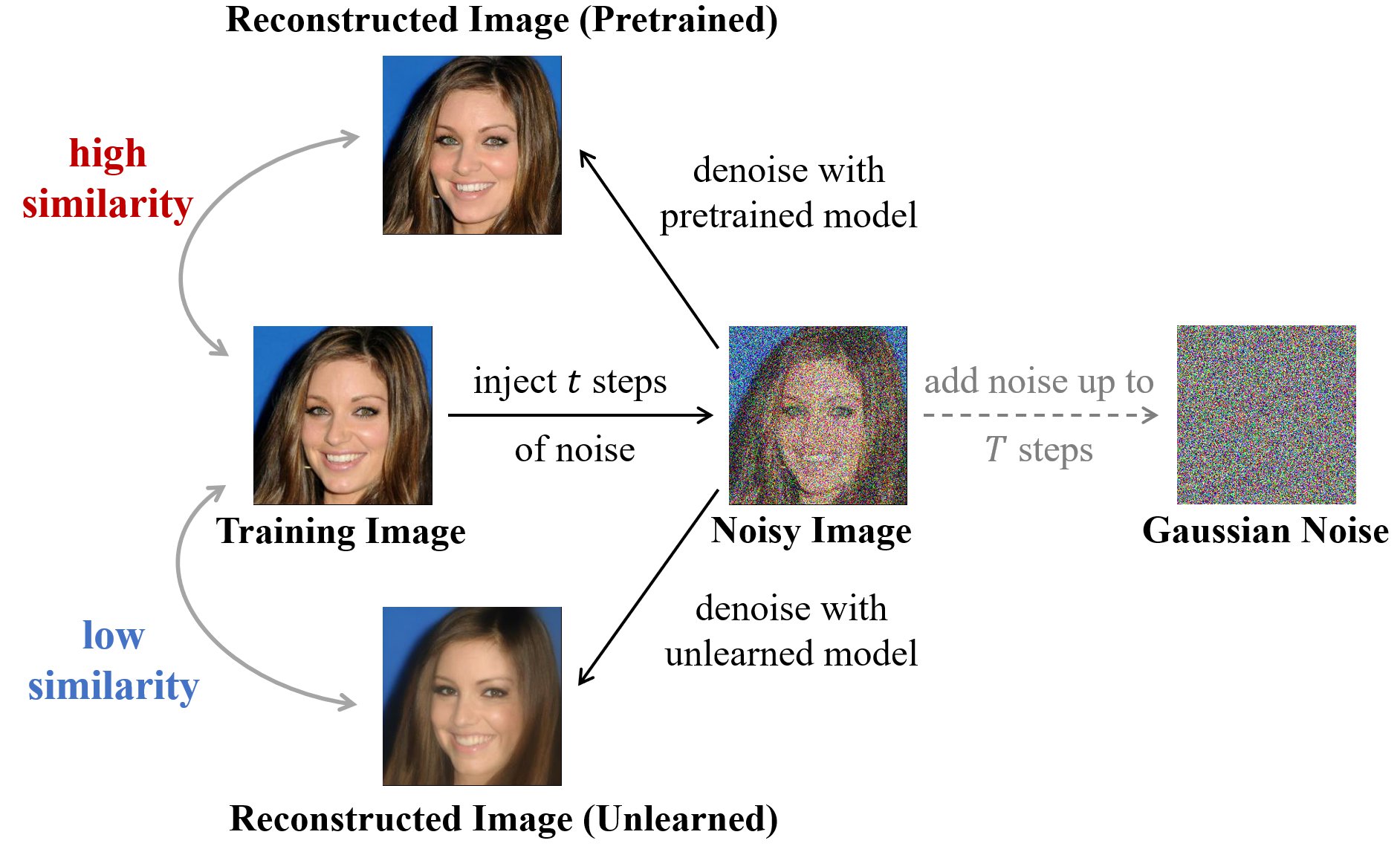}
\caption{
Schematic of the SSCD metric calculation method on CelebA-HQ and CIFAR-10. 
Instead of adding noise up to the full \(T\) steps and generating image from the pure Gaussian noise, we inject noise for only \(t<T\) steps and use the diffusion model to denoise and reconstruct the clean image. The SSCD is then calculated to measure the similarity between the original training image and the reconstructed image, thereby indicating how much of the training image's information is retained by the model.
}
\label{pic:sscd}
\end{figure}

For unlearning metrics, we use Frequency, NLL, and SSCD \cite{sscd} to quantify the extent to which the model memorizes a given image. The details are as follows: 
for the MNIST T‑Shirt dataset, because it is trivial to distinguish whether images sampled from the unlearned model belong to T‑shirts or handwritten digits, we directly assess unlearning performance by calculating the frequency of T-Shirt in the generated images. 
Following the procedure in \citet{song}, we calculate the negative log-likelihood (NLL) to quantify the model’s memorization strength on the specific images.
For CelebA-HQ and CIFAR-10, following \citet{siss}, instead of generating images from the random Gaussian noise, we inject \(t\) steps of noise into the clean image and then use the unlearned model to denoise and reconstruct the clean image. We then compute SSCD between the original training image and the reconstructed image to measure the memorization strength of the unlearned model for this image, as shown in Figure \ref{pic:sscd}.
For the Stable Diffusion dataset, due to the limited number of images sampled during each evaluation, we directly calculate the average SSCD between the images sampled by the unlearned model and the original training image to evaluate the unlearning performance.

\subsubsection{Implementation details.}
For the MNIST T-Shirt dataset constructed by \citet{siss}, we directly employ the pretrained model provided by them. The pretrained models for CelebA‑HQ and CIFAR‑10 are adopted from \citet{ddpm}. For Stable Diffusion, we use v1.4 for experiments.
The noise injection intensity for SSCD calculation is set to \(t=T/4\), consistent with \citet{siss}.
In our experiments, unless otherwise stated, we set the hyperparameter \(k\) in our method to \(10\) and adjust \(\lambda\) in each dataset so that the unlearning term and the regularization term are of similar order of magnitude. The hyperparameters for the other methods are set according to their original works.
Since Stable Diffusion v1.4 is a latent diffusion model that performs the diffusion process in the latent space, in this experiment our method searches for \(k\)-nearest neighbors in the latent space to ensure consistency with the theoretical analysis. For all other experiments with non-latent diffusion models, the \(k\)-nearest neighbors are defined in the pixel domain.

\subsection{Main Results}
This section presents the metric evaluations of each method on each dataset. 
For clarity, method names in the tables are formatted as follows: \textit{italic} for pretrained models, \textbf{bold} for methods that preserve generation quality, and plain font for methods that severely degrade generation quality. The best result among the quality-preserving methods is also highlighted in \textbf{bold}.

\subsubsection{MNIST T-Shirt.}
We conduct experiments under 10 different random seeds and report both the mean and standard deviation of the performance. After fine-tuning for 50 steps, we sample 50,000 images and evaluate the extent of unlearning achieved by each method. The results are presented in Table \ref{tab:mnist_tshirt}. Among all methods that preserve generation quality, our method achieves the strongest unlearning performance.

\begin{table*}[!ht]
\centering
\caption{Results on MNIST T-Shirt}
\label{tab:mnist_tshirt}
\begin{tabular*}{0.8\textwidth}{@{\hspace{1.5em}\extracolsep{\fill}} c|cc|cc @{\hspace{1.5em}}}
\toprule
\multirow{2}{*}{\textbf{Method}} & \multicolumn{2}{c|}{Unlearning} & \multicolumn{2}{c}{Quality} \\
& \textbf{Frequency}\(\downarrow\) & \textbf{NLL}\(\uparrow\) & \textbf{FID}\(\downarrow\) & \textbf{IS}\(\uparrow\) \\
\midrule
\textit{Pretrained} & 0.8252\% {\scriptsize $\pm$ 0.0010\%} & 0.81 {\scriptsize $\pm$ 0.05} & 1.17 {\scriptsize $\pm$ 0.00} & 9.67 {\scriptsize $\pm$ 0.00} \\
\midrule
NegGrad & 0.0000\% {\scriptsize $\pm$ 0.0000\%} & 14.84 {\scriptsize $\pm$ 1.04} & 276.65 {\scriptsize $\pm$ 14.35} & 6.60 {\scriptsize $\pm$ 0.33} \\
EraseDiff & 0.0004\% {\scriptsize $\pm$ 0.0012\%} & 8.77 {\scriptsize $\pm$ 0.29} & 6.92 {\scriptsize $\pm$ 5.48} & 9.14 {\scriptsize $\pm$ 0.36} \\
\midrule
\textbf{Vanilla} & 0.3598\% {\scriptsize $\pm$ 0.2933\%} & 1.14 {\scriptsize $\pm$ 0.09} & 2.16 {\scriptsize $\pm$ 0.72} & \textbf{9.52} {\scriptsize $\pm$ 0.09} \\
\textbf{SISS} & 0.0530\% {\scriptsize $\pm$ 0.0747\%} & 4.53 {\scriptsize $\pm$ 0.50} & 3.05 {\scriptsize $\pm$ 1.58} & 9.37 {\scriptsize $\pm$ 0.24} \\
\textbf{Ours} & \textbf{0.0000\%} {\scriptsize $\pm$ 0.0000\%} & \textbf{8.28} {\scriptsize $\pm$ 0.03} & \textbf{2.09} {\scriptsize $\pm$ 1.04} & 9.48 {\scriptsize $\pm$ 0.06} \\
\bottomrule
\end{tabular*}
\end{table*}

\subsubsection{CelebA-HQ.}
We randomly select 10 facial images from the CelebA-HQ dataset, and separately unlearn those chosen images. After fine-tuning for 40 steps, 10,000 images are sampled to evaluate the generation quality. The average unlearning performance and standard deviations are shown in Table \ref{tab:celeba_hq}. The experimental results show that, among all methods that preserve generation quality, our method yields the best evaluation metrics.

\begin{table*}[!ht]
\centering
\caption{Results on CelebA-HQ}
\label{tab:celeba_hq}
\begin{tabular*}{0.6\textwidth}{@{\hspace{1.5em}\extracolsep{\fill}} c|cc|c @{\hspace{1.5em}}}
\toprule
\multirow{2}{*}{\textbf{Method}} & \multicolumn{2}{c|}{Unlearning} & \multicolumn{1}{c}{Quality} \\
& \textbf{NLL}\(\uparrow\) & \textbf{SSCD}\(\downarrow\) & \textbf{FID}\(\downarrow\) \\
\midrule
\textit{Pretrained} & 1.29 {\scriptsize $\pm$ 0.13} & 0.88 {\scriptsize $\pm$ 0.02} & 17.99 {\scriptsize $\pm$ 0.00} \\
\midrule
NegGrad & 6.93 {\scriptsize $\pm$ 1.12} & 0.14 {\scriptsize $\pm$ 0.07} & 424.62 {\scriptsize $\pm$ 47.04} \\
EraseDiff & 2.59 {\scriptsize $\pm$ 0.26} & 0.33 {\scriptsize $\pm$ 0.05} & 113.15 {\scriptsize $\pm$ 19.32} \\
\midrule
\textbf{Vanilla} & 1.28 {\scriptsize $\pm$ 0.13} & 0.89 {\scriptsize $\pm$ 0.03} & 22.20 {\scriptsize $\pm$ 2.24} \\
\textbf{SISS} & 1.44 {\scriptsize $\pm$ 0.27} & 0.50 {\scriptsize $\pm$ 0.11} & 22.02 {\scriptsize $\pm$ 1.64} \\
\textbf{Ours} & \textbf{1.51} {\scriptsize $\pm$ 0.25} & \textbf{0.41} {\scriptsize $\pm$ 0.16} & \textbf{21.26} {\scriptsize $\pm$ 2.59} \\
\bottomrule
\end{tabular*}
\end{table*}

\subsubsection{CIFAR-10.}
We randomly choose 10 images from the CIFAR-10 dataset and separately conduct experiments on them. We sample 10,000 images for generation quality evaluation after 60 fine-tuning steps. The average results are reported in Table \ref{tab:cifar10}. The results indicate that our method outperforms all other methods that preserve generation quality.

\begin{table*}[!ht]
\centering
\caption{Results on CIFAR-10}
\label{tab:cifar10}
\begin{tabular*}{0.7\textwidth}{@{\hspace{1.5em}\extracolsep{\fill}} c|cc|cc @{\hspace{1.5em}}}
\toprule
\multirow{2}{*}{\textbf{Method}} & \multicolumn{2}{c|}{Unlearning} & \multicolumn{2}{c}{Quality} \\
& \textbf{NLL}\(\uparrow\) & \textbf{SSCD}\(\downarrow\) & \textbf{FID}\(\downarrow\) & \textbf{IS}\(\uparrow\) \\
\midrule
\textit{Pretrained} & 3.22 {\scriptsize $\pm$ 0.36} & 0.54 {\scriptsize $\pm$ 0.06} & 5.27 {\scriptsize $\pm$ 0.00} & 9.29 {\scriptsize $\pm$ 0.00} \\
\midrule
NegGrad & 4.33 {\scriptsize $\pm$ 0.35} & 0.23 {\scriptsize $\pm$ 0.06} & 81.02 {\scriptsize $\pm$ 33.19} & 6.32 {\scriptsize $\pm$ 1.59} \\
EraseDiff & 3.55 {\scriptsize $\pm$ 0.30} & 0.48 {\scriptsize $\pm$ 0.09} & 29.22 {\scriptsize $\pm$ 8.68} & 8.32 {\scriptsize $\pm$ 0.45} \\
\midrule
\textbf{Vanilla} & 3.22 {\scriptsize $\pm$ 0.36} & 0.55 {\scriptsize $\pm$ 0.08} & 8.77 {\scriptsize $\pm$ 0.00} & 9.14 {\scriptsize $\pm$ 0.00} \\
\textbf{SISS} & 3.21 {\scriptsize $\pm$ 0.37} & 0.45 {\scriptsize $\pm$ 0.06} & 9.05 {\scriptsize $\pm$ 0.64} & 9.29 {\scriptsize $\pm$ 0.13} \\
\textbf{Ours} & \textbf{3.44} {\scriptsize $\pm$ 0.32} & \textbf{0.37} {\scriptsize $\pm$ 0.05} & \textbf{8.45} {\scriptsize $\pm$ 0.42} & \textbf{9.42} {\scriptsize $\pm$ 0.14} \\
\bottomrule
\end{tabular*}
\end{table*}

\subsubsection{Stable Diffusion.}
\begin{figure}[t]
\centering
\includegraphics[width=0.98\columnwidth]{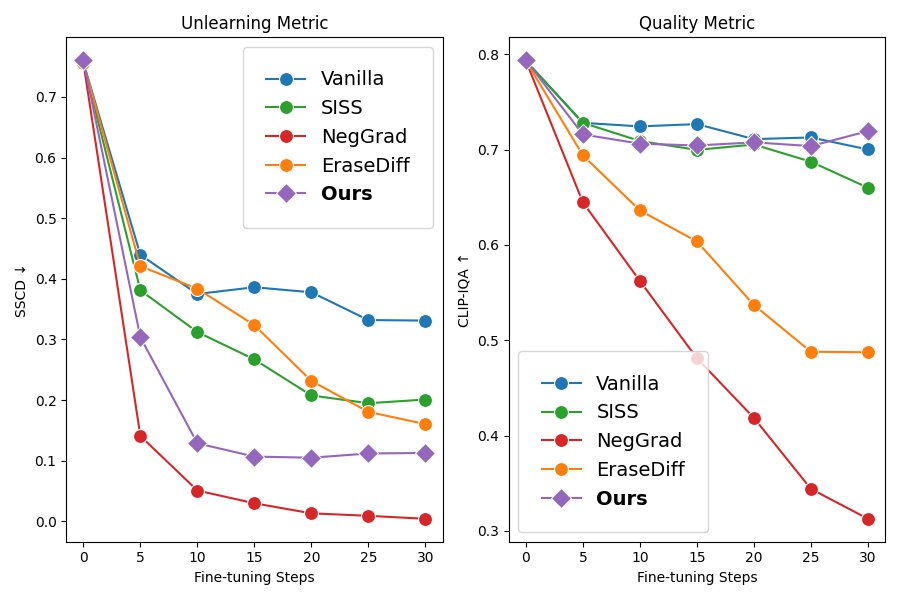}
\caption{Results on Stable Diffusion.}
\label{pic:sd_results}
\end{figure}

We conducted experiments independently on 48 text-image pairs.
For each experiment, a total of 30 steps of fine-tuning are performed, with an evaluation every 5 steps. 
For each evaluation, 16 images are sampled to compute the generation quality metric CLIP‑IQA and the unlearning metric SSCD.
The averaged results are shown in Figure \ref{pic:sd_results}.
The results show that the unlearning metric of our method is only slightly lower than that of NegGrad method, which causes serious degradation of the generation quality, implying that our method is the best among all quality-preserving methods.

\subsection{Ablation Study}
In the ablation study, we discuss two important parts in our method: the choice of hyperparameter \(k\) and the effect of the regularization term. All experiments conducted in this section are performed on the MNIST T-Shirt dataset.

\begin{table*}[!ht]
\centering
\caption{Ablation study on the choice of hyperparameter \(k\)}
\label{tab:ablation_K}
\begin{tabular*}{0.8\textwidth}{@{\hspace{1.5em}\extracolsep{\fill}} c|cc|cc @{\hspace{1.5em}}}
\toprule
\multirow{2}{*}{\(\boldsymbol{k}\)} & \multicolumn{2}{c|}{Unlearning} & \multicolumn{2}{c}{Quality} \\
& \textbf{Frequency}\(\downarrow\) & \textbf{NLL}\(\uparrow\) & \textbf{FID}\(\downarrow\) & \textbf{IS}\(\uparrow\) \\
\midrule
\textbf{1} & 0.0000\% {\scriptsize $\pm$ 0.0000\%} & 8.29 {\scriptsize $\pm$ 0.04} & 2.22 {\scriptsize $\pm$ 1.02} & 9.48 {\scriptsize $\pm$ 0.06} \\
\textbf{10} & 0.0000\% {\scriptsize $\pm$ 0.0000\%} & 8.28 {\scriptsize $\pm$ 0.03} & 2.09 {\scriptsize $\pm$ 1.04} & 9.48 {\scriptsize $\pm$ 0.06} \\
\textbf{100} & 0.0000\% {\scriptsize $\pm$ 0.0000\%} & 8.28 {\scriptsize $\pm$ 0.04} & 2.04 {\scriptsize $\pm$ 1.02} & 9.48 {\scriptsize $\pm$ 0.06} \\
\textbf{1000} & 0.0000\% {\scriptsize $\pm$ 0.0000\%} & 8.28 {\scriptsize $\pm$ 0.04} & 2.01 {\scriptsize $\pm$ 1.05} & 9.49 {\scriptsize $\pm$ 0.07} \\
\midrule
\textit{Pretrained} & 0.8252\% {\scriptsize $\pm$ 0.0010\%} & 0.81 {\scriptsize $\pm$ 0.05} & 1.17 {\scriptsize $\pm$ 0.00} & 9.67 {\scriptsize $\pm$ 0.00} \\
\bottomrule
\end{tabular*}
\end{table*}

\begin{table*}[!ht]
\centering
\caption{Ablation study on the effect of the regularization term.}
\label{tab:ablation_remain}
\begin{tabular*}{0.8\textwidth}{@{\hspace{1.5em}\extracolsep{\fill}} c|cc|cc @{\hspace{1.5em}}}
\toprule
\multirow{2}{*}{\textbf{Setting}} & \multicolumn{2}{c|}{Unlearning} & \multicolumn{2}{c}{Quality} \\
& \textbf{Frequency}\(\downarrow\) & \textbf{NLL}\(\uparrow\) & \textbf{FID}\(\downarrow\) & \textbf{IS}\(\uparrow\) \\
\midrule
\textbf{w/ reg.} & 0.0000\% {\scriptsize $\pm$ 0.0000\%} & 8.28 {\scriptsize $\pm$ 0.03} & 2.09 {\scriptsize $\pm$ 1.04} & 9.48 {\scriptsize $\pm$ 0.06} \\
\textbf{w/o reg.} & 0.0000\% {\scriptsize $\pm$ 0.0000\%} & 9.74 {\scriptsize $\pm$ 0.09} & 14.22 {\scriptsize $\pm$ 3.81} & 9.40 {\scriptsize $\pm$ 0.08} \\
\midrule
\textit{Pretrained} & 0.8252\% {\scriptsize $\pm$ 0.0010\%} & 0.81 {\scriptsize $\pm$ 0.05} & 1.17 {\scriptsize $\pm$ 0.00} & 9.67 {\scriptsize $\pm$ 0.00} \\
\bottomrule
\end{tabular*}
\end{table*}

\subsubsection{Choice of hyperparameter k.}
We keep all other experimental settings fixed and vary the hyperparameter \(k\) in our method. The corresponding results are presented in Table \ref{tab:ablation_K}. As \(k\) increases, the unlearning metrics remains essentially unchanged, while the quality metrics exhibits a slight improvement. Therefore, to strike a balance between the performance of our method and the implementation simplicity of the loss function, we recommend setting \(k = 10\) based on the empirical observations.

\subsubsection{Effect of the regularization term.}
To demonstrate the necessity of incorporating the regularization term in the loss function of our method, we hold all other experimental settings fixed and compare the performance of two variants of our loss function: one including both the unlearning term and the regularization term, and the other retaining only the unlearning term. The results, summarized in Table \ref{tab:ablation_remain}, reveal that although omitting the regularization term yields a marginal improvement in the unlearning metrics, it also causes a pronounced degradation in generation quality. This deterioration indicates that the absence of the regularization term induces over-forgetting, thereby validating the crucial role of the regularization term in preserving the model’s generative quality.

\section{CONCLUSION}

In this paper, we propose ReTrack, a novel data unlearning method for diffusion models that redirects the denoising trajectories toward the \(k\)-nearest neighbors. Using importance sampling, we first derive an unbiased loss that concentrates the fine‑tuning procedure on the regions surrounding the unlearning set. Then, we approximate this loss by keeping only the \(k\) largest terms, enabling fast unlearning while preserving generation quality. Across MNIST T-Shirt, CelebA-HQ, CIFAR-10, and Stable Diffusion, ReTrack achieves the best trade-off between unlearning strength and generation quality compared with prior methods.

\section*{Acknowledgements}
 The authors are with the Department of Electronic Engineering, Beijing National Research Center for Information Science and Technology, Tsinghua University, Beijing 100084, China. This work was supported by the National Key Research and Development Program of China (Grant No. 2025YFF0515601) and the National Natural Science Foundation of China (NSAF U2230201).

\bibliographystyle{aaai2026}
\bibliography{refs}

%%%%%%%%%%%%%%%%%%%%%%%%%%%%%%%%%%%%%%%%%%%%%%%%%%%%%%%%%%%%
\section*{Checklist}

\begin{enumerate}

  \item For all models and algorithms presented, check if you include:
  \begin{enumerate}
    \item A clear description of the mathematical setting, assumptions, algorithm, and/or model. [Yes]
    \item An analysis of the properties and complexity (time, space, sample size) of any algorithm. [Yes]
    \item (Optional) Anonymized source code, with specification of all dependencies, including external libraries. [Yes]
  \end{enumerate}

  \item For any theoretical claim, check if you include:
  \begin{enumerate}
    \item Statements of the full set of assumptions of all theoretical results. [Yes]
    \item Complete proofs of all theoretical results. [Yes]
    \item Clear explanations of any assumptions. [Yes]     
  \end{enumerate}

  \item For all figures and tables that present empirical results, check if you include:
  \begin{enumerate}
    \item The code, data, and instructions needed to reproduce the main experimental results (either in the supplemental material or as a URL). [Yes]
    \item All the training details (e.g., data splits, hyperparameters, how they were chosen). [Yes]
    \item A clear definition of the specific measure or statistics and error bars (e.g., with respect to the random seed after running experiments multiple times). [Yes]
    \item A description of the computing infrastructure used. (e.g., type of GPUs, internal cluster, or cloud provider). [Yes]
  \end{enumerate}

  \item If you are using existing assets (e.g., code, data, models) or curating/releasing new assets, check if you include:
  \begin{enumerate}
    \item Citations of the creator If your work uses existing assets. [Yes]
    \item The license information of the assets, if applicable. [Yes]
    \item New assets either in the supplemental material or as a URL, if applicable. [Yes]
    \item Information about consent from data providers/curators. [Not Applicable]
    \item Discussion of sensible content if applicable, e.g., personally identifiable information or offensive content. [Not Applicable]
  \end{enumerate}

  \item If you used crowdsourcing or conducted research with human subjects, check if you include:
  \begin{enumerate}
    \item The full text of instructions given to participants and screenshots. [Not Applicable]
    \item Descriptions of potential participant risks, with links to Institutional Review Board (IRB) approvals if applicable. [Not Applicable]
    \item The estimated hourly wage paid to participants and the total amount spent on participant compensation. [Not Applicable]
  \end{enumerate}

\end{enumerate}

\clearpage
\appendix
\thispagestyle{empty}

% Supplementary material: To improve readability, you must use a single-column format for the supplementary material.
\onecolumn
\aistatstitle{Supplementary Materials}

\section{PROOFS OF PROPOSITIONS}

In this section, we provide the detailed proofs of the propositions stated in the methodology section of the main text.

\subsection{Proof of Proposition~\ref{prop:equivalence}}
\equivalence*
\begin{proof}
Denoting the discrete uniform distributions on \(A_r\) and \(A_u\) as
\[
\begin{aligned}
q_{A_r}(\boldsymbol{a}) &= \frac{1}{|A_r|}\,\mathbf{1}_{A_r}(\boldsymbol{a}),\\
q_{A_u}(\boldsymbol{a}) &= \frac{1}{|A_u|}\,\mathbf{1}_{A_u}(\boldsymbol{a}),
\end{aligned}
\]
where \(|\cdot|\) denotes the cardinality of a finite set, and the indicator function is defined as
\[
\mathbf{1}_{A}(\boldsymbol{a}) =
\begin{cases}
1, & \boldsymbol{a} \in A,\\
0, & \boldsymbol{a} \notin A.
\end{cases}
\]
Expanding the expectations over \(\boldsymbol{a}_r\) and \(\boldsymbol{x}_t\) in \(\mathcal{L}_{\text{vanilla}}\), we have
\begin{align*}
\mathcal{L}_{\text{vanilla}}(\boldsymbol{\theta}) &= \mathbb{E}_{t,\boldsymbol{a}_r\sim A_r,\boldsymbol{x}_t\sim q_t(\cdot|\boldsymbol{a}_r)}\left[\|\boldsymbol{\epsilon}_{\boldsymbol{\theta}}(\boldsymbol{x}_t,t)-\boldsymbol{\epsilon}\|_2^2\right] \\
&= \mathbb{E}_{t}\left[\sum_{\boldsymbol{a}_r\in A_r}\int_{\mathbb{R}^d} q_{A_r}(\boldsymbol{a}_r)q_t(\boldsymbol{x}_t|\boldsymbol{a}_r)  \left\|\boldsymbol{\epsilon}_{\boldsymbol{\theta}}(\boldsymbol{x}_t,t)-\frac{\boldsymbol{x}_t-\gamma_t \boldsymbol{a}_r}{\sigma_t}\right\|_2^2  \mathrm{d}\boldsymbol{x}_t\right] \\
&= \mathbb{E}_{t}\left[\sum_{\boldsymbol{a}_r\in A_r}\int_{\mathbb{R}^d} q_t(\boldsymbol{x}_t)\frac{q_{A_r}(\boldsymbol{a}_r)q_t(\boldsymbol{x}_t|\boldsymbol{a}_r)}{q_t(\boldsymbol{x}_t)}  \left\|\boldsymbol{\epsilon}_{\boldsymbol{\theta}}(\boldsymbol{x}_t,t)-\frac{\boldsymbol{x}_t-\gamma_t \boldsymbol{a}_r}{\sigma_t}\right\|_2^2  \mathrm{d}\boldsymbol{x}_t\right].
\end{align*}
By defining the weights as
\begin{align*}
w_t(\boldsymbol{x}_t|\boldsymbol{a}_r) &= \frac{q_{A_r}(\boldsymbol{a}_r)q_t(\boldsymbol{x}_t|\boldsymbol{a}_r)}{q_t(\boldsymbol{x}_t)} \\
&= \frac{{q_{A_r}(\boldsymbol{a}_r)}q_t(\boldsymbol{x}_t|\boldsymbol{a}_r)}{\sum_{\boldsymbol{a}_r'\in A_r}{q_{A_r}(\boldsymbol{a}_r')}q_t(\boldsymbol{x}_t|\boldsymbol{a}_r')} \\
&= \frac{q_t(\boldsymbol{x}_t|\boldsymbol{a}_r)}{\sum_{\boldsymbol{a}_r'\in A_r}q_t(\boldsymbol{x}_t|\boldsymbol{a}_r')}
\end{align*}
and expanding the first \(q_t(\boldsymbol{x}_t)\) in the integral using the law of total probability
\begin{align*}
q_t(\boldsymbol{x}_t) = \sum_{\boldsymbol{a}_u\in A_u}q_{A_u}(\boldsymbol{a}_u)q_t(\boldsymbol{x}_t|\boldsymbol{a}_u),
\end{align*}
 we obtain
\begin{align*}
\mathcal{L}_{\text{vanilla}}(\boldsymbol{\theta}) &= \mathbb{E}_{t}\left[\sum_{\boldsymbol{a}_r\in A_r}\int_{\mathbb{R}^d} \sum_{\boldsymbol{a}_u\in A_u}q_{A_u}(\boldsymbol{a}_u)q_t(\boldsymbol{x}_t|\boldsymbol{a}_u)  
w_t(\boldsymbol{x}_t|\boldsymbol{a}_r)
\left\|\boldsymbol{\epsilon}_{\boldsymbol{\theta}}(\boldsymbol{x}_t,t)-\frac{\boldsymbol{x}_t-\gamma_t \boldsymbol{a}_r}{\sigma_t}\right\|_2^2  \mathrm{d}\boldsymbol{x}_t\right] \\
&= \mathbb{E}_{t}\left[\sum_{\boldsymbol{a}_u\in A_u}\int_{\mathbb{R}^d}
q_{A_u}(\boldsymbol{a}_u) q_t(\boldsymbol{x}_t|\boldsymbol{a}_u)  
\sum_{\boldsymbol{a}_r\in A_r}w_t(\boldsymbol{x}_t|\boldsymbol{a}_r)
\left\|\boldsymbol{\epsilon}_{\boldsymbol{\theta}}(\boldsymbol{x}_t,t)-\frac{\boldsymbol{x}_t-\gamma_t \boldsymbol{a}_r}{\sigma_t}\right\|_2^2  \mathrm{d}\boldsymbol{x}_t\right] \\
&= \mathbb{E}_{t,\boldsymbol{a}_u\sim A_u,\boldsymbol{x}_t\sim q_t(\cdot|\boldsymbol{a}_u)}   \left[\sum_{\boldsymbol{a}_r\in A_r} w_t(\boldsymbol{x}_t|\boldsymbol{a}_r)\left\|\boldsymbol{\epsilon}_{\boldsymbol{\theta}}(\boldsymbol{x}_t,t)-\frac{\boldsymbol{x}_t-\gamma_t \boldsymbol{a}_r}{\sigma_t}\right\|_2^2\right] \\
&= \mathcal{L}_{\text{unlearn}}(\boldsymbol{\theta}). 
\end{align*}
\end{proof}

\subsection{Proof of Proposition~\ref{prop:maxk}}
\maxk*
\begin{proof}
Define the auxiliary variable
\[
\boldsymbol{z}_t = \frac{\boldsymbol{x}_t - \gamma_t \boldsymbol{a}_r}{\sigma_t}.
\]
Since \(\boldsymbol{x}_t\sim q_t(\cdot|\boldsymbol{a}_u)\), we have
\[
\boldsymbol{z}_t \sim \mathcal{N}(\cdot;\boldsymbol{\mu}_t,\mathbf{I}_d),
\]
where the expectation of the Gaussian distribution is
\[
\boldsymbol{\mu}_t = \frac{\gamma_t}{\sigma_t}(\boldsymbol{a}_u-\boldsymbol{a}_r).
\]
Then, the expected values of the likelihoods are given by
\begin{align*}
\mathbb{E}_{t,\boldsymbol{x}_t\sim q_t(\cdot|\boldsymbol{a}_u)}\left[q_t(\boldsymbol{x}_t|\boldsymbol{a}_r)\right] 
=& \mathbb{E}_{t,\boldsymbol{x}_t\sim q_t(\cdot|\boldsymbol{a}_u)}\left[\frac{1}{(2\pi\sigma_t^2)^{\frac d2}}\exp\left(-\frac{\|\boldsymbol{x}_t-\gamma_t \boldsymbol{a}_r\|_2^2}{2\sigma_t^2}\right)\right] \\
=& \mathbb{E}_{t,\boldsymbol{z}_t \sim \mathcal{N}(\cdot;\boldsymbol{\mu}_t,\mathbf{I}_d)}\left[\frac{1}{(2\pi\sigma_t^2)^{\frac d2}}\exp\left(-\frac{1}{2}\|\boldsymbol{z}_t\|_2^2\right)\right] \\
\propto& \sum_{t=1}^{T} \sigma_t^{-d} \int_{\mathbb{R}^d} \exp\left(-\frac{1}{2}\|\boldsymbol{z}_t-\boldsymbol{\mu}_t\|_2^2\right)\exp\left(-\frac{1}{2}\|\boldsymbol{z}_t\|_2^2\right) \mathrm{d}\boldsymbol{z}_t \\
=& \sum_{t=1}^{T} \sigma_t^{-d} \int_{\mathbb{R}^d} \exp\left(-\|\boldsymbol{z}_t\|_2^2+\boldsymbol{\mu}_t^\top\boldsymbol{z}_t-\frac{1}{2}\|\boldsymbol{\mu}_t\|_2^2\right) \mathrm{d}\boldsymbol{z}_t \\
=& \sum_{t=1}^{T} \sigma_t^{-d} \exp\left(-\frac{1}{4}\|\boldsymbol{\mu}_t\|_2^2\right)\int_{\mathbb{R}^d} \exp\left(-\left\|\boldsymbol{z}_t-\frac{\boldsymbol{\mu}_t}{2}\right\|_2^2\right) \mathrm{d}\boldsymbol{z}_t \\
\propto& \sum_{t=1}^{T} \sigma_t^{-d} \exp\left(-\frac{1}{4}\|\boldsymbol{\mu}_t\|_2^2\right) \\
=& \sum_{t=1}^{T} \sigma_t^{-d} \exp\left(-\frac{\gamma_t^2}{4\sigma_t^2}\|\boldsymbol{a}_u-\boldsymbol{a}_r\|_2^2\right), 
\end{align*}
so maximizing the likelihoods \(q_t(\boldsymbol{x}_t|\boldsymbol{a}_r)\) in expectation is equivalent to minimizing the Euclidean distance \(\|\boldsymbol{a}_u - \boldsymbol{a}_r\|_2\). Consequently, the \(k\) samples in \(A_r\) with the largest expected likelihoods are the \(k\)-nearest neighbors of \(\boldsymbol{a}_u\) under the Euclidean metric.
\end{proof}

\section{COMPARISON WITH PRIOR METHOD}
\label{sec:comparison_prior}

While our proposed ReTrack and the recent SISS \cite{siss} both utilize importance sampling for data unlearning in diffusion models, they differ fundamentally in their theoretical formulations and the intrinsic role of importance sampling.

According to the discussion in Appendix A.1 (Stability Analysis and Interpretation of SISS) of the SISS paper, their unlearning objective is equivalent to a vanilla loss and a heavily biased gradient-ascent term:
\[ \mathcal{L}_{\text{SISS}}(\boldsymbol{\theta}) = \mathcal{L}_{\text{vanilla}}(\boldsymbol{\theta}) + s \mathcal{L}_{\text{NegGrad}}(\boldsymbol{\theta}) \]
where \( s \) is a coefficient chosen to dynamically balance the gradient norms. In their method, importance sampling acts primarily as a computational trick to unify the forward-pass computation of these two distinct loss terms, effectively halving the number of function evaluations. As implied by their own ablation studies, importance sampling is not the core driver of unlearning in SISS; the method fundamentally relies on the biased \( \mathcal{L}_{\text{NegGrad}} \) component.

In contrast, our unlearning objective is theoretically unbiased and is formulated as a combination of an unlearning term and a regularization term:
\[ \mathcal{L}_{\text{ReTrack}}(\boldsymbol{\theta}) = \lambda \mathcal{L}_{\text{unlearn}}(\boldsymbol{\theta}) + (1-\lambda) \mathcal{L}_{\text{vanilla}}(\boldsymbol{\theta}) \]
We identify that while the original vanilla loss perfectly preserves generation quality due to its unbiasedness, it suffers from extremely low sampling efficiency for unlearning. Consequently, ReTrack employs importance sampling specifically to shift the sampling distribution of the vanilla loss toward the unlearning set, yielding a mathematically equivalent but vastly more sample-efficient unlearning loss, \( \mathcal{L}_{\text{unlearn}} \). In our framework, importance sampling is intrinsic: removing it would reduce our approach back to the vanilla loss, eliminating any meaningful unlearning effect.

Finally, we note that the numerical discrepancies between the results reported in our evaluation and those in the SISS publication are simply due to the different numbers of fine-tuning steps utilized in our experimental setup.

\section{DETAILED EXPERIMENTAL SETUP}

In this section, we present a detailed description of the experimental setup, including the hardware and software environment, the datasets and models, the evaluation metrics, and the training hyperparameter settings employed in the experiments.

\subsection{Computational Environment}
All experiments were conducted on NVIDIA A100‑SXM4‑40GB GPU hardware and implemented using the \texttt{PyTorch} framework and the Hugging Face \texttt{Diffusers} library. The AdamW optimizer \cite{adamw} was employed for network optimization.

\subsection{Datasets and Models}
We conducted experiments on four datasets: MNIST T‑Shirt, CelebA‑HQ, CIFAR‑10, and Stable Diffusion. For CelebA‑HQ and CIFAR‑10, we employed the pretrained models provided in \citet{ddpm}. The MNIST T‑Shirt dataset and its pretrained model were adopted directly from \citet{siss}. For Stable Diffusion, we used version 1.4 and the corresponding dataset constructed by \citet{siss}. All diffusion models use UNet as the backbone, with a total diffusion steps \(T=1000\). We use the EMA version for all pretrained models. 

During fine‑tuning, all models were initialized from the provided pretrained checkpoints without any additional modifications. For Stable Diffusion model, we trained only the denoising UNet while keeping all other components fixed.

\subsection{Evaluation Metrics}
For the Frequency metric on MNIST T‑Shirt, we used the \(\ell_2\) distance to determine whether a generated image belongs to T-Shirts. We computed the NLL metric according to the process presented in \citet{song}. The SSCD metric was computed using the default \texttt{sscd\_disc\_mixup} model recommended by \citet{sscd}. FID, IS, and CLIP‑IQA were all evaluated via the \texttt{torchmetrics} library. For IS, we used a ten‑class classifier trained on MNIST for the MNIST T‑Shirt dataset and the standard Inception V3 classifier for CIFAR‑10.

\subsection{Training Hyperparameter Settings}
On the MNIST T‑Shirt, CelebA‑HQ, CIFAR‑10, and Stable Diffusion datasets, the learning rate during fine‑tuning was kept constant at \(5\times10^{‑5}\), \(5\times10^{‑6}\), \(2\times10^{‑5}\), and \(1\times10^{‑5}\) respectively.
These values were chosen to ensure good performance of the vanilla method.
The training batch size was set to 128, 64, 128, and 16 respectively.
For the MNIST T‑Shirt, CelebA‑HQ, and CIFAR‑10 datasets, we used AdamW with beta parameters set to (0.95, 0.999) and weight decay of \(1\times10^{‑6}\).
For the Stable Diffusion dataset, beta was set to (0.9, 0.999) and weight decay to \(1\times10^{‑2}\).
To balance generation quality with inference efficiency, we set the inference steps to 1000 for CIFAR‑10 and to 50 for all other datasets.
All hyperparameter configurations are available in the YAML files under the \texttt{/configs} directory in our code.

\section{ADDITIONAL RESULTS}

\begin{figure*}[t]
  \centering
  \includegraphics[width=\textwidth]{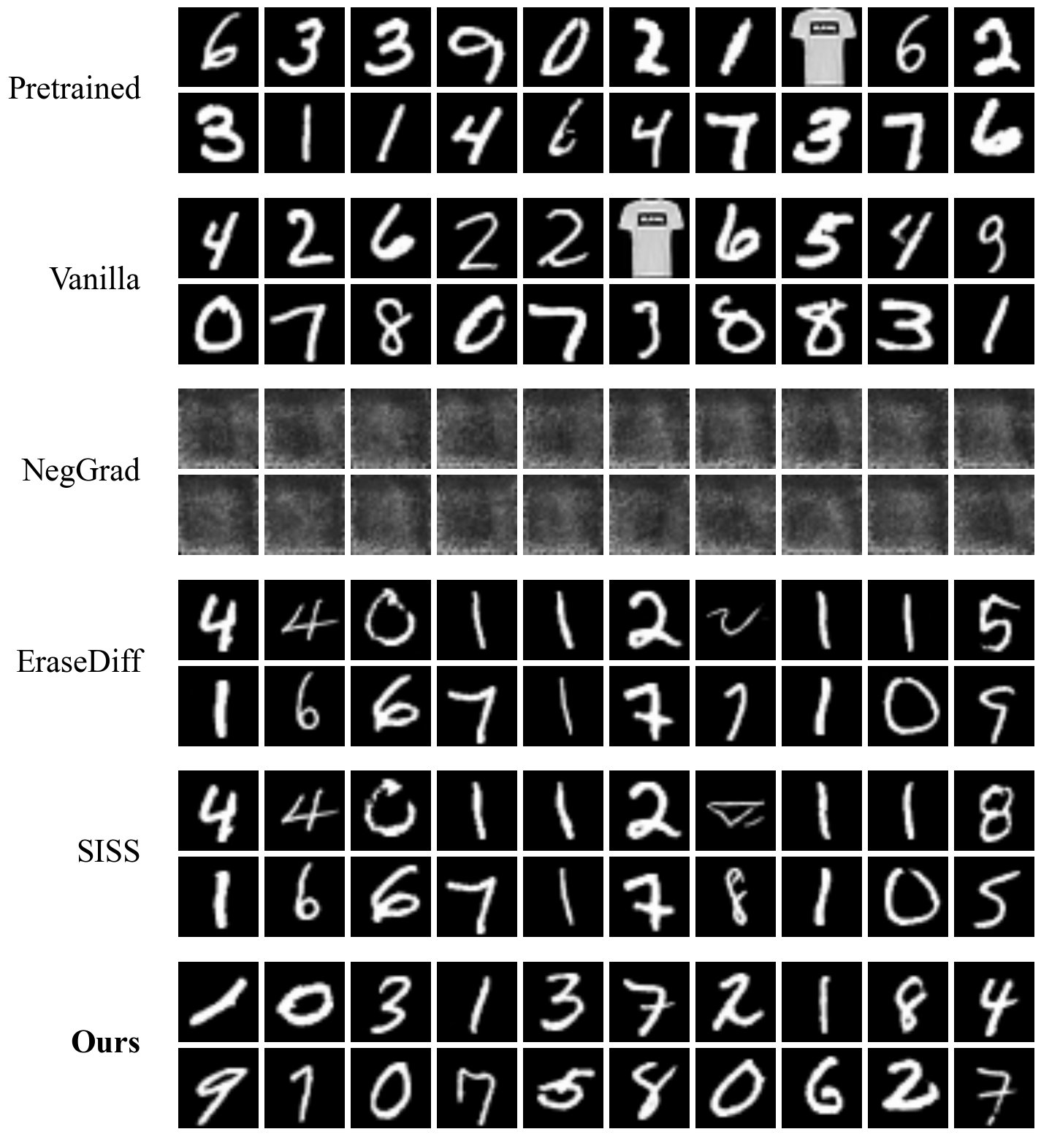}
  \caption{Partial sampling results from pretrained and unlearned models on the MNIST T-Shirt dataset.}
  \label{pic:mnist}
\end{figure*}

\begin{figure*}[t]
  \centering
  \includegraphics[width=\textwidth]{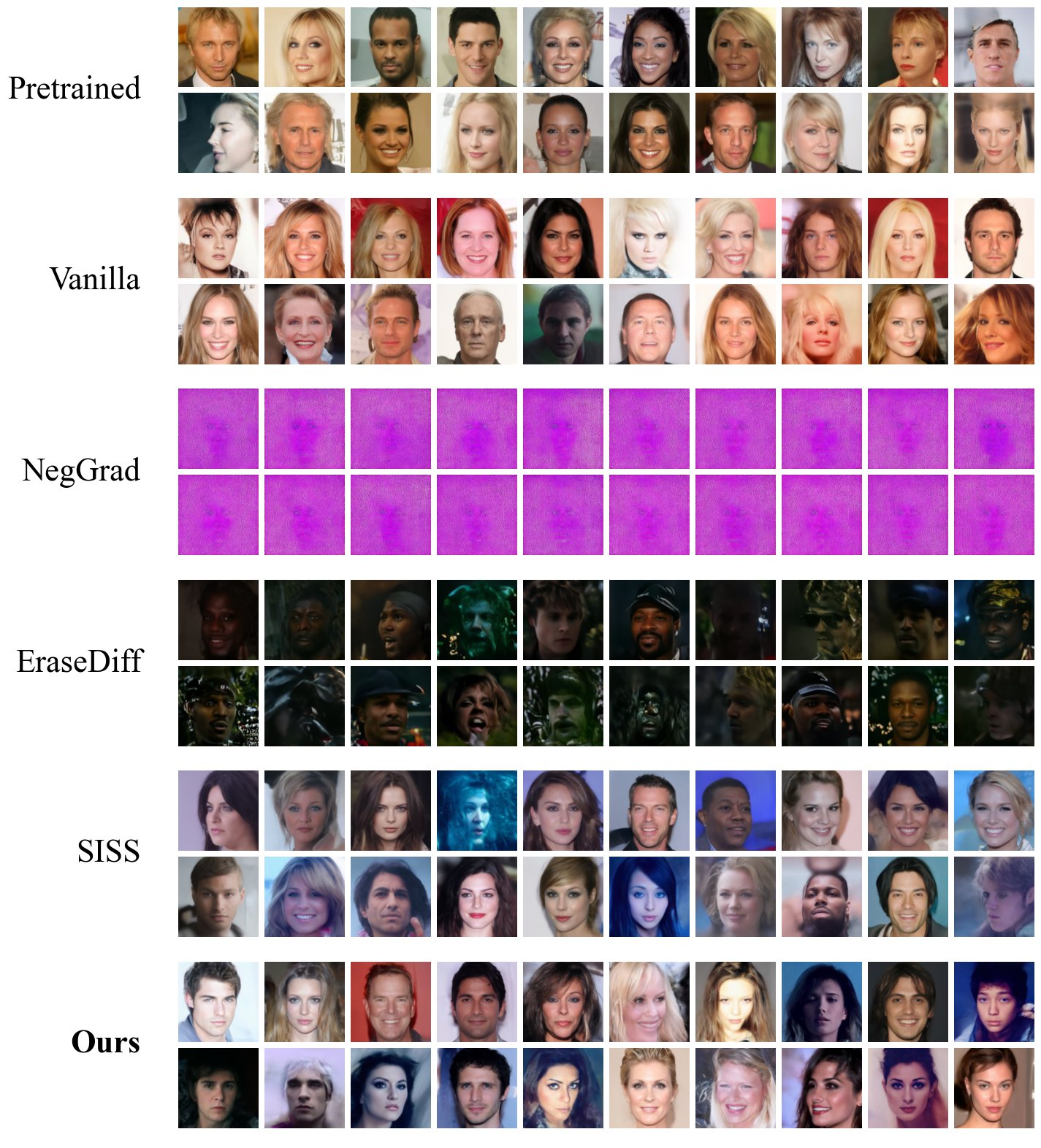}
  \caption{Partial sampling results from pretrained and unlearned models on the CelebA-HQ dataset.}
  \label{pic:celeba}
\end{figure*}

\begin{figure*}[t]
  \centering
  \includegraphics[width=\textwidth]{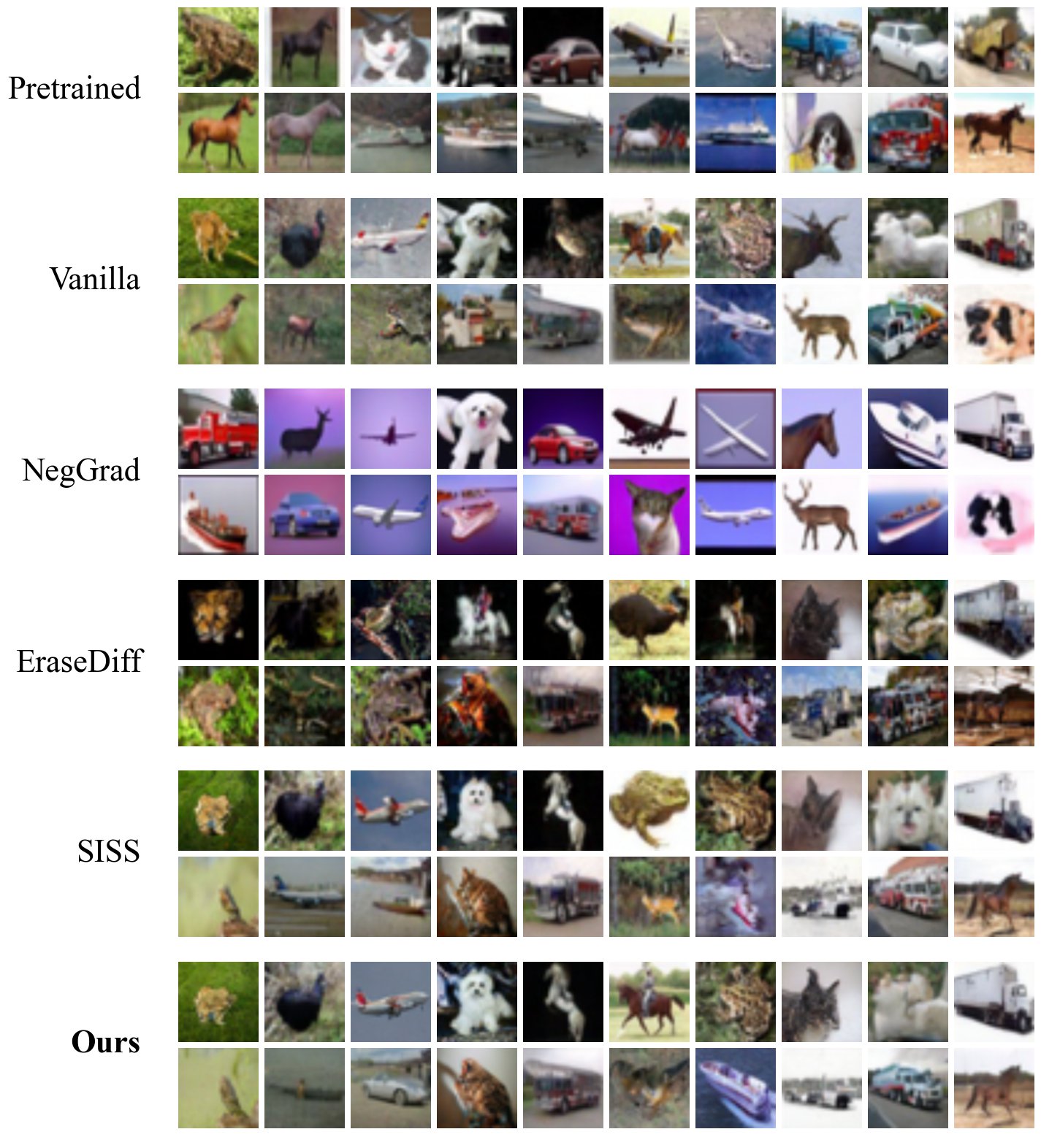}
  \caption{Partial sampling results from pretrained and unlearned models on the CIFAR-10 dataset.}
  \label{pic:cifar}
\end{figure*}

\begin{figure*}[t]
  \centering
  \includegraphics[width=0.9\textwidth]{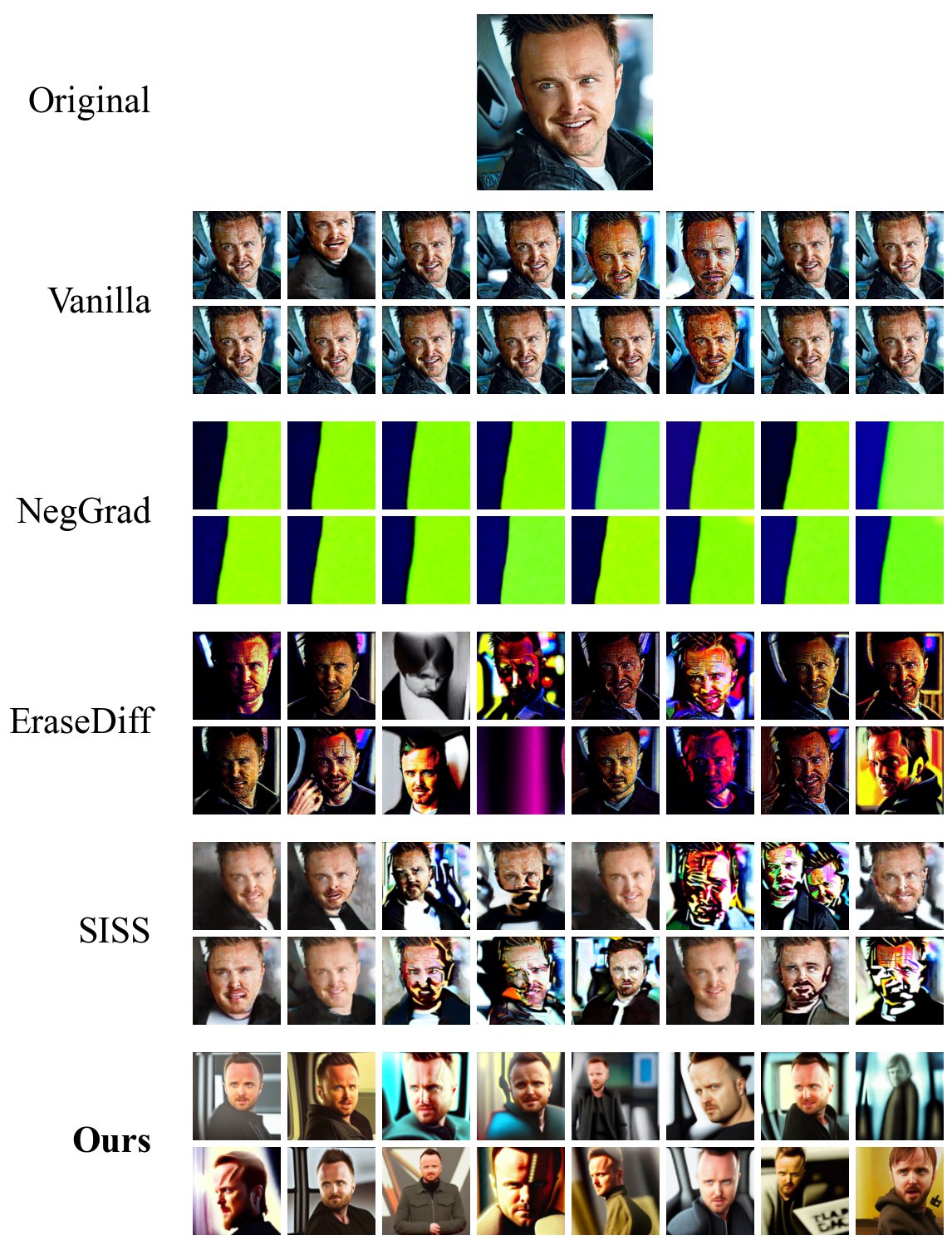}
  \caption{Sampling results from pretrained and unlearned models on the Stable Diffusion dataset conditioned on the prompt ``Aaron Paul to Play Luke Skywalker at LACMA Reading of \textit{The Empire Strikes Back}".}
  \label{pic:aaron}
\end{figure*}

\begin{figure*}[t]
  \centering
  \includegraphics[width=0.9\textwidth]{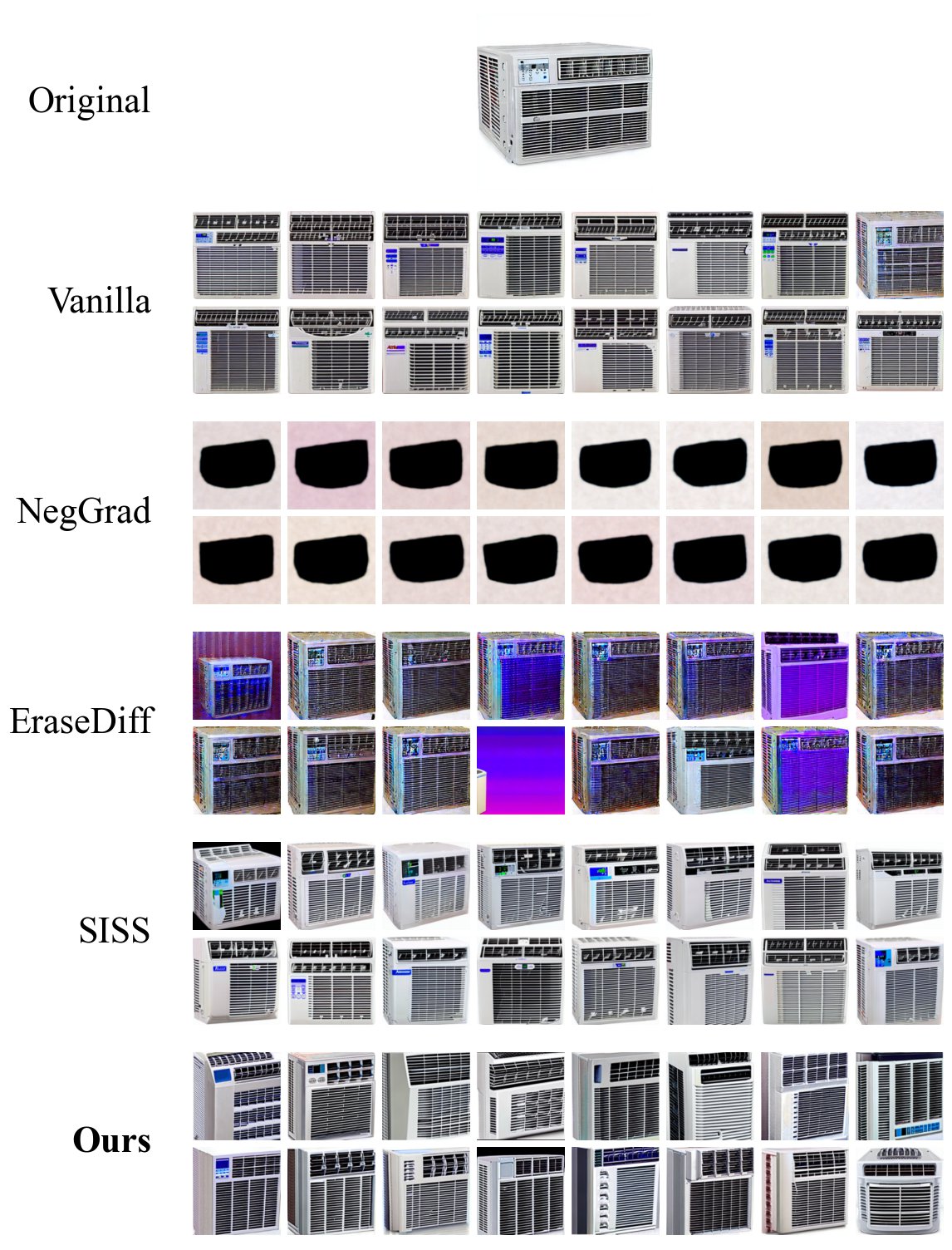}
  \caption{Sampling results from pretrained and unlearned models on the Stable Diffusion dataset conditioned on the prompt ``Air Conditioners \& Parts".}
  \label{pic:air}
\end{figure*}

\begin{figure*}[t]
  \centering
  \includegraphics[width=0.9\textwidth]{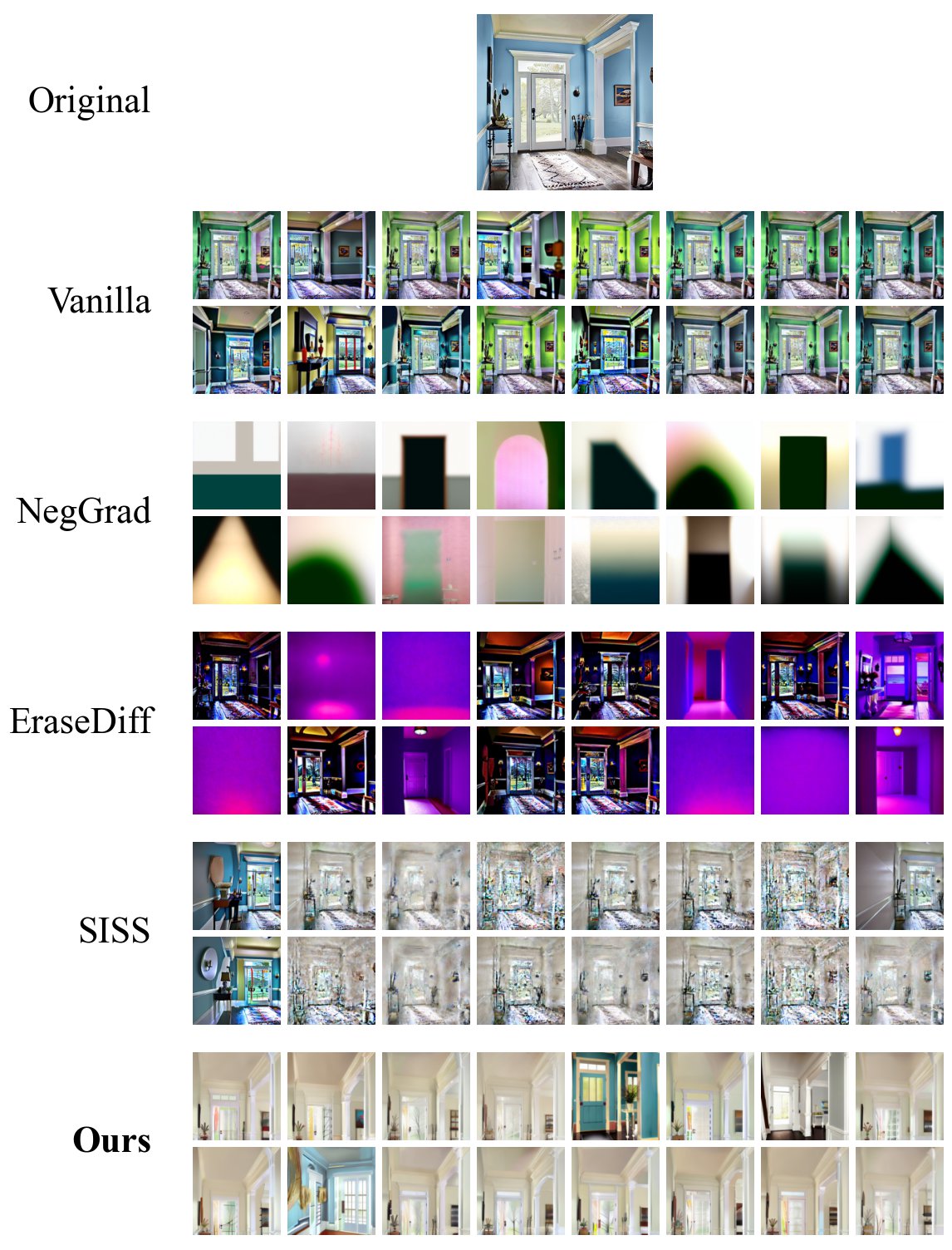}
  \caption{Sampling results from pretrained and unlearned models on the Stable Diffusion dataset conditioned on the prompt ``Foyer painted in SALTY AIR".}
  \label{pic:foyer}
\end{figure*}

\begin{figure*}[t]
  \centering
  \includegraphics[width=\textwidth]{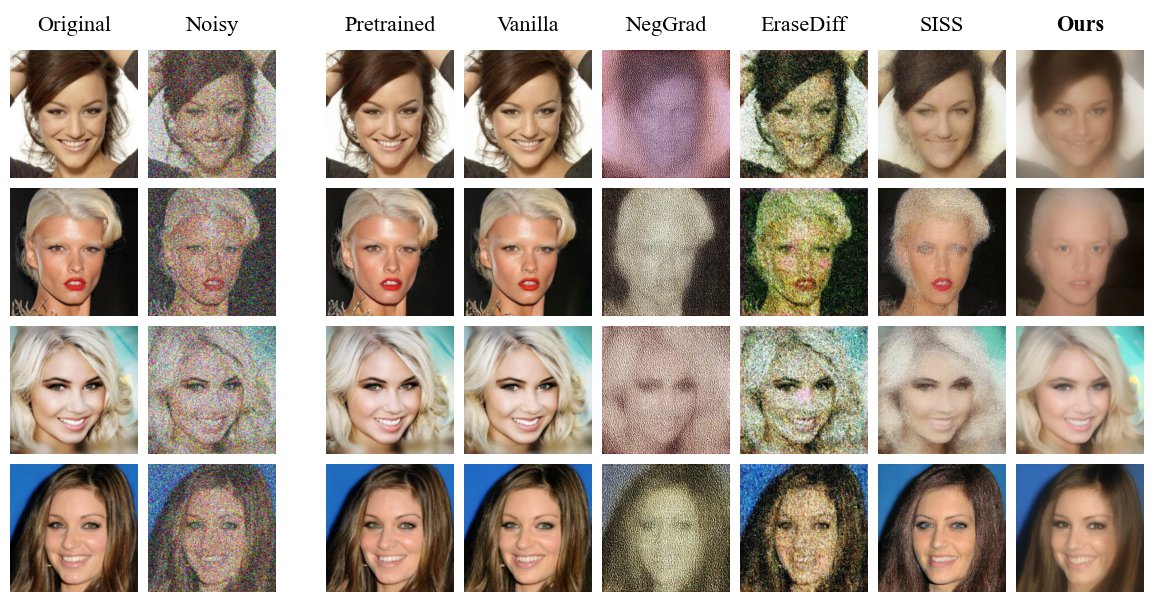}
  \caption{Partial reconstruction images from pretrained and unlearned models on the CelebA-HQ dataset when computing the SSCD metric.}
  \label{pic:celeba_siss}
\end{figure*}

\begin{figure*}[t]
  \centering
  \includegraphics[width=\textwidth]{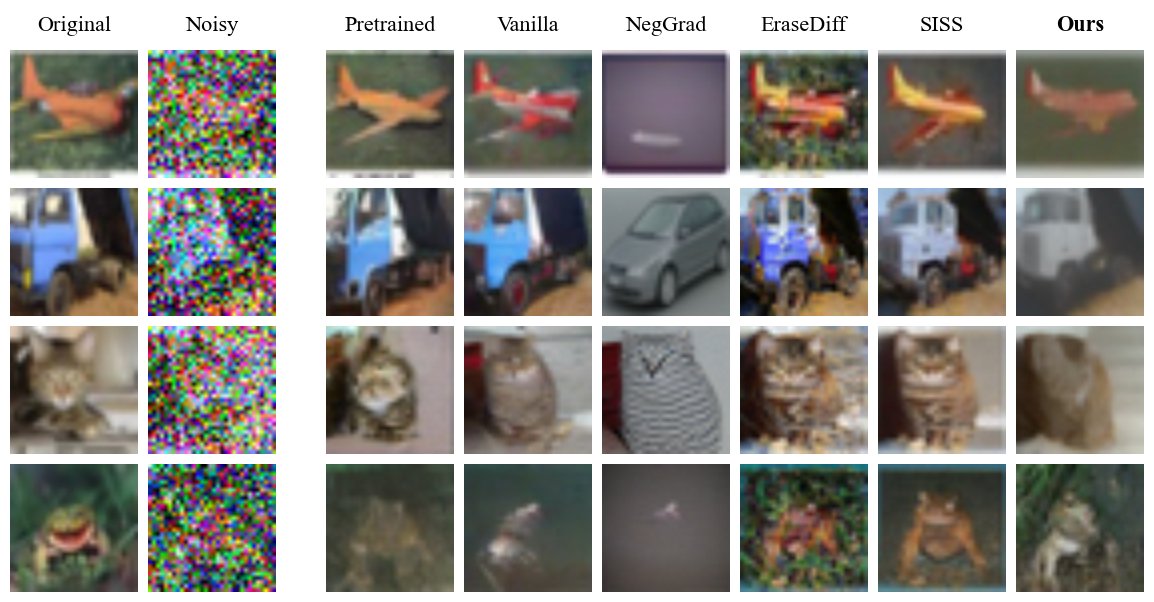}
  \caption{Partial reconstruction images from pretrained and unlearned models on the CIFAR-10 dataset when computing the SSCD metric.}
  \label{pic:cifar_siss}
\end{figure*}

In this section, we present more visualization results on each dataset.
For image generation quality visualizations on each dataset, see Figure \ref{pic:mnist}, \ref{pic:celeba}, \ref{pic:cifar}, \ref{pic:aaron}, \ref{pic:air} and \ref{pic:foyer}.
For the images used during the calculation of the SSCD metric for the CelebA-HQ and CIFAR-10 datasets, see Figure \ref{pic:celeba_siss} and \ref{pic:cifar_siss}.

% \bibliography{refs}

\end{document}